\documentclass[final,5p,times,twocolumn]{elsarticle}

\usepackage{amsmath, amsfonts, amssymb}

\usepackage{graphicx}
\usepackage[dvipsnames, table]{xcolor}  
\usepackage{caption}
\usepackage{subcaption}

\usepackage{booktabs}
\usepackage{multirow}
\usepackage{makecell}
\usepackage{tabularx}
\usepackage{threeparttable}
\usepackage{colortbl}
\usepackage{array}
\usepackage{dcolumn}

\usepackage{algorithm}
\usepackage{algorithmic}
\usepackage{listings}
\usepackage{verbatim}

\usepackage{stfloats}
\usepackage{placeins}
\usepackage{wrapfig}
\usepackage{adjustbox}
\usepackage{microtype}

\usepackage[normalem]{ulem}  

\usepackage{pifont}
\usepackage{tcolorbox}

\usepackage[hidelinks]{hyperref}

\usepackage{array,multirow}
\newcolumntype{P}[1]{>{\centering\arraybackslash}m{#1}}

\begin{document}

\begin{frontmatter}




\title{A Survey of Large Language Models for Data Challenges in Graphs}

\author[1]{Mengran~Li}
\ead{limr39@mail2.sysu.edu.cn}

\author[2]{Pengyu~Zhang}%
\ead{p.zhang@uva.nl}

\author[1]{Wenbin~Xing}
\ead{xingwb@mail2.sysu.edu.cn}

\author[2]{Yijia~Zheng}
\ead{y.zheng@uva.nl}

\author[3]{Klim Zaporojets}
\ead{klim@cs.au.dk}

\author[1]{Junzhou~Chen}
\ead{chenjunzhou@mail.sysu.edu.cn}

\author[1]{Ronghui~Zhang\corref{cor1}%
}
\ead{zhangrh25@mail.sysu.edu.cn}

\author[4]{Yong~Zhang}
\ead{zhangyong2010@bjut.edu.cn}

\author[5]{Siyuan~Gong}
\ead{sgong@chd.edu.cn}

\author[6]{Jia~Hu}
\ead{hujia@tongji.edu.cn}

\author[7]{Xiaolei~Ma}
\ead{xiaolei@buaa.edu.cn}

\author[8]{Zhiyuan~Liu}
\ead{zhiyuanl@seu.edu.cn}

\author[2]{Paul Groth}
\ead{p.t.groth@uva.nl}

\author[2]{Marcel Worring}
\ead{m.worring@uva.nl}

\cortext[cor1]{Corresponding author.}

\affiliation[1]{organization={Guangdong Key Laboratory of Intelligent Transportation System, School of Intelligent Systems Engineering, Shenzhen Campus of Sun Yat-sen University},
                city={Shenzhen},
                postcode={518107}, 
                state={Guangdong},
                country={China}}

\affiliation[2]{organization={University of Amsterdam}, 
                city={Amsterdam},
                country={The Netherlands}}

\affiliation[3]{organization={Aarhus University},
              city={Aarhus},
              country={Denmark}}
              
\affiliation[4]{organization={Beijing Institute of Artificial Intelligence, Beijing University of Technology},
                city={Beijing},
                postcode={100124}, 
                country={China}}
                
\affiliation[5]{organization={School of Information and Engineering, Chang’an University},
                city={Xi'an},
                postcode={710064}, 
                state={Shaanxi}, 
                country={China}}
                
\affiliation[6]{organization={Key Laboratory of Road and Traffic Engineering of the Ministry of Education, Tongji University},
                city={Shanghai},
                postcode={201804}, 
                country={China}}
                
\affiliation[7]{organization={Key Laboratory of Intelligent Transportation Technology and System, School of Transportation Science and Engineering, Beihang University},
                city={Beijing},
                postcode={100191}, 
                country={China}}
                
\affiliation[8]{organization={Jiangsu Key Laboratory of Urban ITS, Jiangsu Province Collaborative Innovation Center of Modern Urban Traffic Technologies, School of Transportation, Southeast University},
                city={Nanjing},
                postcode={210096}, 
                state={Jiangsu},
                country={China}}

\begin{abstract}
Graphs are a widely used paradigm for representing non-Euclidean data, with applications ranging from social network analysis to biomolecular prediction. While graph learning has achieved remarkable progress, real-world graph data presents a number of challenges that significantly hinder the learning process. In this survey, we focus on four fundamental data-centric challenges: \textbf{(1) Incompleteness}, real-world graphs have missing nodes, edges, or attributes; \textbf{(2) Imbalance}, the distribution of the labels of nodes or edges and their structures for real-world graphs are highly skewed; \textbf{(3) Cross-domain Heterogeneity}, graphs from different domains exhibit incompatible feature spaces or structural patterns; and \textbf{(4) Dynamic Instability}, graphs evolve over time in unpredictable ways. Recently, Large Language Models (LLMs) offer the potential to tackle these challenges by leveraging rich semantic reasoning and external knowledge. This survey focuses on how LLMs can address four fundamental data-centric challenges in graph-structured data, thereby improving the effectiveness of graph learning. For each challenge, we review both traditional solutions and modern LLM-driven approaches, highlighting how LLMs contribute unique advantages. Finally, we discuss open research questions and promising future directions in this emerging interdisciplinary field. To support further exploration, we have curated a repository of recent advances on graph learning challenges: \url{https://github.com/limengran98/Awesome-Literature-Graph-Learning-Challenges}.  
\end{abstract}




\end{frontmatter}

\section{Introduction}\label{s1}

Graphs provide a flexible and expressive framework for modeling complex systems composed of interconnected entities. Unlike grid-structured data such as images or sequences, graphs encode non-Euclidean topologies with irregular neighborhood structures and varying connectivity patterns. To support learning on such structures, graph learning aims to derive low-dimensional node or graph-level representations that preserve both structural dependencies and semantic information from the original graph \cite{xia2021graph, khoshraftar2024survey, ju2024comprehensive, DBLP:journals/tnn/LiZWHY25}. Due to these representational advantages, graph learning has been successfully applied across a wide range of domains, including social network analysis \cite{yang2024graph}, personalized recommendation \cite{wang2023deep}, transportation optimization \cite{rahmani2023graph}, financial modeling \cite{motie2024financial}, and bioinformatics \cite{yang2024poisoning, xie2025pseudo}. As a result, it has become a foundational technique for extracting knowledge from relational data and enabling models to reason over structured domains.

Despite this progress, real-world graph data often exhibits complexities, including sparsity, noise, domain divergence, and dynamic evolution, introducing practical difficulties for graph learning \cite{ju2024survey}. For instance, in social networks, cold-start users may lack sufficient links or profile features, making it difficult to learn reliable representations \cite{wang2023deep}. In fraud detection, the data is often highly imbalanced, where fraudulent behavior accounts for only a small fraction of all activities, and labels are sparse and noisy \cite{rajput2022temporal}. In smart cities, traffic systems involve heterogeneous and dynamically evolving road networks, posing difficulties for models to adapt to changing topologies in real time \cite{rahmani2023graph, 10797699, zhang2025mmstflownet}. As summarized in Figure~\ref{fig1}, these concrete scenarios highlight four \textbf{fundamental data-centric challenges} in real-world graphs:

\begin{figure*}[t]
\centering
\includegraphics[width=0.9\linewidth]{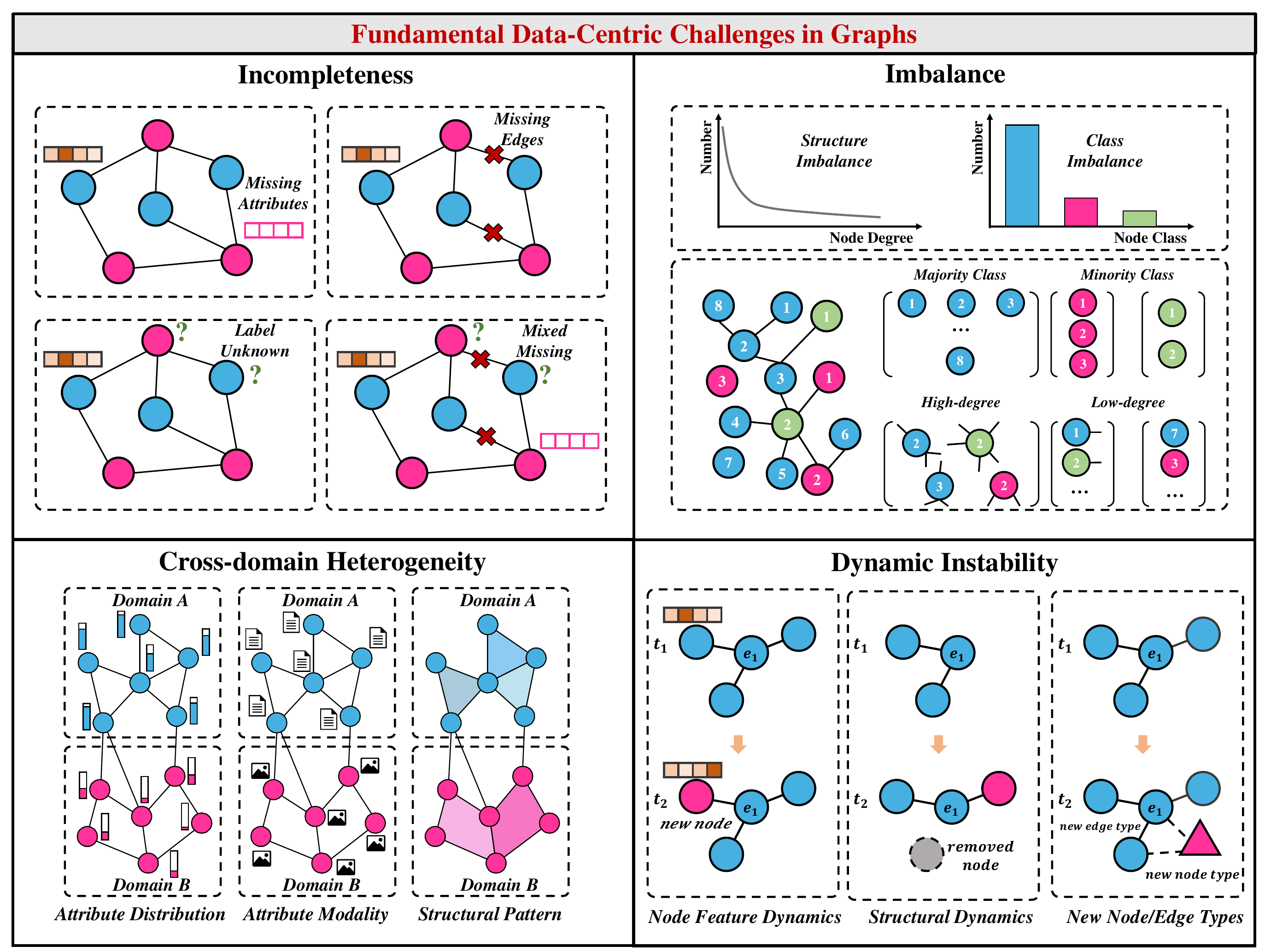}
\caption{The four fundamental challenges emerge of real-world graph complexity: (1) Incompleteness in graphs, where nodes, edges, or attributes are missing, (2) Imbalance in graphs, where the distribution of nodes, edges, or labels is highly skewed, (3) Cross-domain heterogeneity in graphs, where graph data from different domains exhibit semantic and structural discrepancies, and (4) Dynamic instability in graphs, where graphs undergo dynamic changes in topology, attributes, or interactions over time.}
\label{fig1}
\end{figure*}

\textbf{(1) Incompleteness in Graphs}
An incomplete graph refers to a graph where some node or edge information is missing, resulting in an incomplete representation of the structure or attributes that fails to fully reflect the complexity of the real-world system \cite{chen2022learning}. This challenge typically arises from incomplete data collection, missing link information, or outdated knowledge. For example, in social networks, missing user attributes—such as age, location, or interests—can obscure community boundaries, making it difficult to accurately detect user groups or recommend content. Similarly, missing connections between users may conceal important social ties, weakening the structural signals used for clustering or influence analysis. This incompleteness undermines the effectiveness of graph learning.

\textbf{(2) Imbalance in Graphs}
A graph is considered imbalanced when certain categories or subgraphs contain significantly more nodes or edges than others \citep{liu2023survey}. This can lead to some categories dominating the graph, while others are left underrepresented \cite{wang2022imbalanced}. For example, financial transaction networks are characterized by containing a large number of legitimate transactions, and only a small fraction of fraudulent ones \cite{liu2021pick}. Imbalanced training data may cause models to overfit to the majority class, thus impacting the model's ability to generalize and recognize the minority class effectively \cite{ma2023class}. This data imbalance challenge affects many graph learning methods, especially those relying on graph convolutional networks \cite{shi2020multi}.

\textbf{(3) Cross-Domain Heterogeneity in Graphs} 
Cross-domain heterogeneity refers to significant disparities in attributes and structural patterns that arise when graph data are constructed from multiple source domains, especially when these domains have different data modalities or exhibit substantial distribution shifts. For instance, road network data collected from various cities may exhibit considerable heterogeneity, with some cities having grid-like structural patterns, while others exhibit radial configurations \cite{badhrudeen2022geometric}.
This issue becomes even more critical in the rapidly growing field of graph foundation models \cite{liu2025foundation}, which aim to develop generalizable models capable of handling diverse graphs from different domains, such as molecular graphs and citation networks. Cross-domain heterogeneity poses additional challenges for graph learning, as diverse attributes and structural patterns cannot be easily processed in a unified manner \cite{zhang2019dane, yan2024when}. Additionally, domain discrepancies may distort essential semantic and structural signals, complicating the identification of transferable features and limiting the effectiveness of graph learning methods \cite{hassani2022cross, zhu2021shift}. 

\textbf{(4) Dynamic Instability in Graphs}
Dynamic instability in graphs refers to continuous and unpredictable changes in node features and relationships over time. These may include updates to node attributes, alterations in graph structure, or the emergence of new node and edge types. Structural changes often involve adding or removing connections or reorganizing how nodes interact. For example, in a large-scale knowledge graph like Wikidata, an entity initially labeled as an ``athlete'' might later become classified as a ``coach''. This transition entails significant shifts in both node features and their connections within the graph \cite{10.1145/3627673.3679702, DBLP:conf/ecai/ZhangCG24}. Furthermore, these continuous transformations introduce instability, making it difficult for graph learning models to adapt effectively. As a result, performance may degrade in tasks such as node classification, link prediction, and node representation learning \cite{JMLR:v21:19-447}. A deeper understanding of these forms of dynamic instability can facilitate the development of more robust and adaptive graph learning methods, crucial for accurately modeling the evolving real-world graphs.

A variety of graph learning methods have been developed to address these challenges, achieving notable success in specific contexts. For instance, graph completion algorithms \cite{cai2010singular} and graph autoencoders \cite{chen2022learning} have been employed to handle missing data; techniques like GraphSMOTE \cite{zhao2021graphsmote} and re-weighted loss functions \cite{zhang2022dual} are commonly used to mitigate data imbalance; and graph domain adaptation methods, including adversarial regularization \cite{zhang2019dane, shen2020adversarial} and distribution alignment \cite{zhu2021shift, liu2023structural}, have been explored to address cross-domain heterogeneity. Additionally, dynamic graph models such as EvolveGCN \cite{Pareja_Domeniconi_Chen_Ma_Suzumura_Kanezashi_Kaler_Schardl_Leiserson_2020} and DyRep \cite{osti_10190736} have been proposed to model evolving graph structures. While these methods are effective within their respective scopes, they often rely on task-specific designs and require significant domain expertise. 

\begin{figure}[t]
\centering
\includegraphics[width=0.9\linewidth]{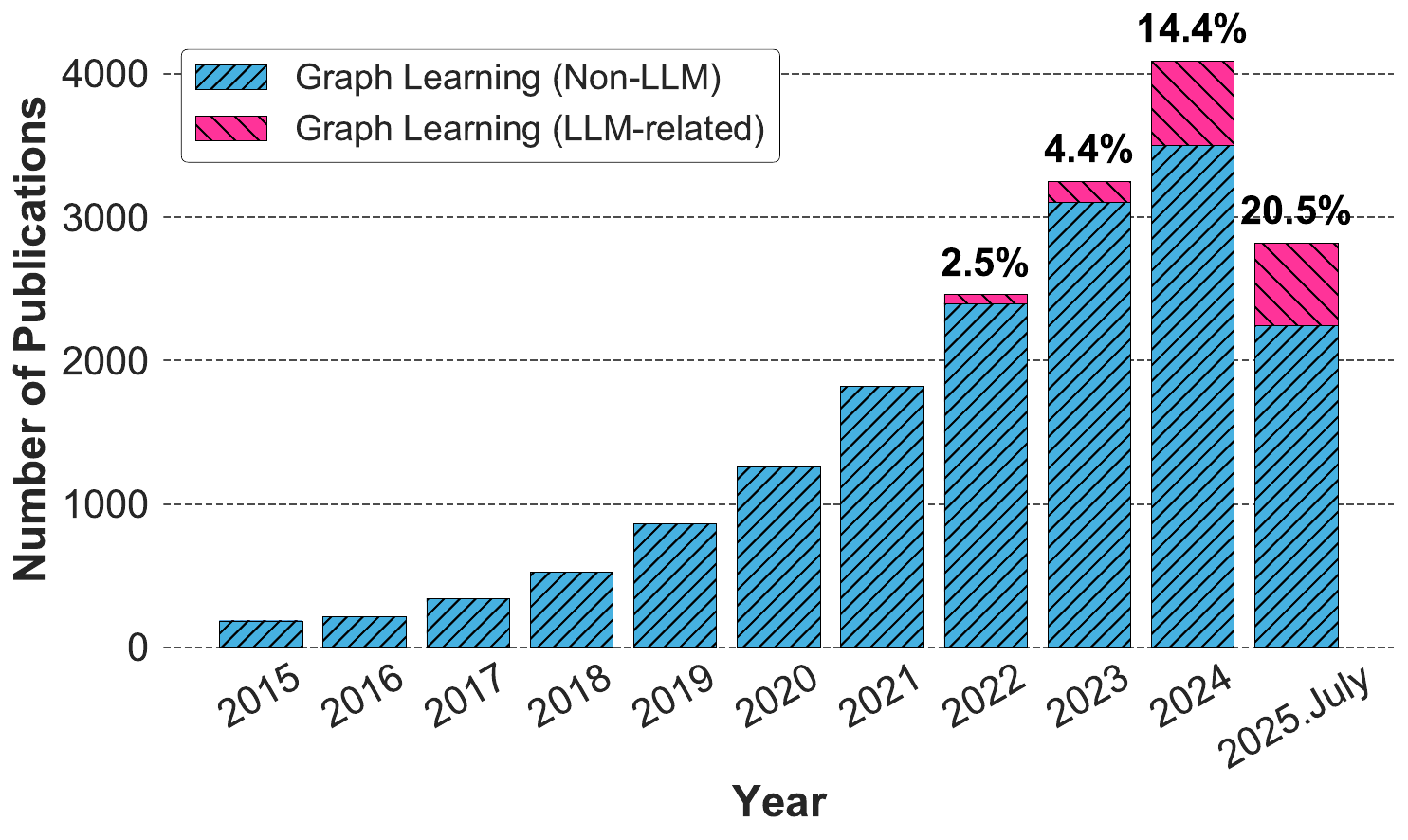}
\caption{Annual publication trends in Graph Learning from January 2015 to July 2025, with a highlighted subset of works related to LLMs. The blue bars represent Graph Learning (non-LLM) publications and the pink segments represent LLM-related publications; the stacked height indicates the total per year. The percentages above each bar denote the proportion of LLM-related papers within the total for that year. Data sourced from Google Scholar (\url{https://scholar.google.com}).}
\label{figst}
\end{figure}

In contrast, the emergence of large language models (LLMs) has opened up new opportunities for mitigating these fundamental challenges \cite{cui2025kllms4rec, jeong2025llm, ong2025dynamic}. With their strong representational capacity, contextual reasoning abilities, and generalization potential, LLMs can extract rich semantic patterns from heterogeneous and noisy data. These capabilities enable LLMs to augment graph data by inferring missing information \cite{brasoveanu2023framing}, synthesizing data for underrepresented classes \cite{yu2025leveraging}, aligning heterogeneous attributes \cite{liu2023evaluating}, and capturing temporal evolution or structural changes in dynamic graphs \cite{margatina-etal-2023-dynamic}. By addressing these issues, LLMs offer a promising complement to traditional graph learning approaches. To reflect this growing interest, Figure~\ref{figst} shows publication trends over the past decade (up to July 2025) related to ``Graph Learning" and ``LLM-based Graph Learning". The statistics indicate that not only has graph learning become an increasingly important topic, but the proportion of research exploring the integration of LLMs into graph learning has also been steadily rising.

This survey provides a comprehensive review of current research at the intersection of graph learning and large language models, centered on the four fundamental challenges introduced above. To ensure broad coverage, we performed an extensive literature search across major databases (arXiv, ACM Digital Library, IEEE Xplore, Elsevier, and Springer, etc.) using keyword combinations such as ``LLMs'', ``graph learning'', ``incompleteness'', ``imbalance'', ``heterogeneity'', ``dynamic graphs''. We focus on recent publications (primarily 2015–2025) in top machine learning, data mining, and NLP venues, as well as relevant preprints. This survey includes papers that meet at least one of the following criteria: (1) the work explicitly addresses one of the four fundamental challenges in graph learning, or (2) the proposed solution incorporates LLMs as a central component. Based on these criteria, we initially collected and reviewed over 1,000 relevant papers. We then categorized them according to the type of challenge addressed and the strategies employed, ultimately selecting a curated set of over 380 representative works. For each challenge, we briefly discuss conventional (non-LLM) techniques to provide background context, while our primary focus is to highlight the novel contributions and advantages introduced by LLM-based approaches.

The structure of the paper is organized as follows: Section \ref{s2} provides an analysis and summary of traditional methods and LLM-based approaches for handling graph learning challenges. Section \ref{s3}, building on the existing body of research, proposes new technical paths and solutions. Section \ref{s4} discusses the potential future directions for graph learning research. Finally, Section \ref{s5} concludes the paper.

\section{Overview of Graph Learning and LLMs}\label{s2}

\begin{figure}[t]
\centerline{\includegraphics[width=0.9\linewidth]{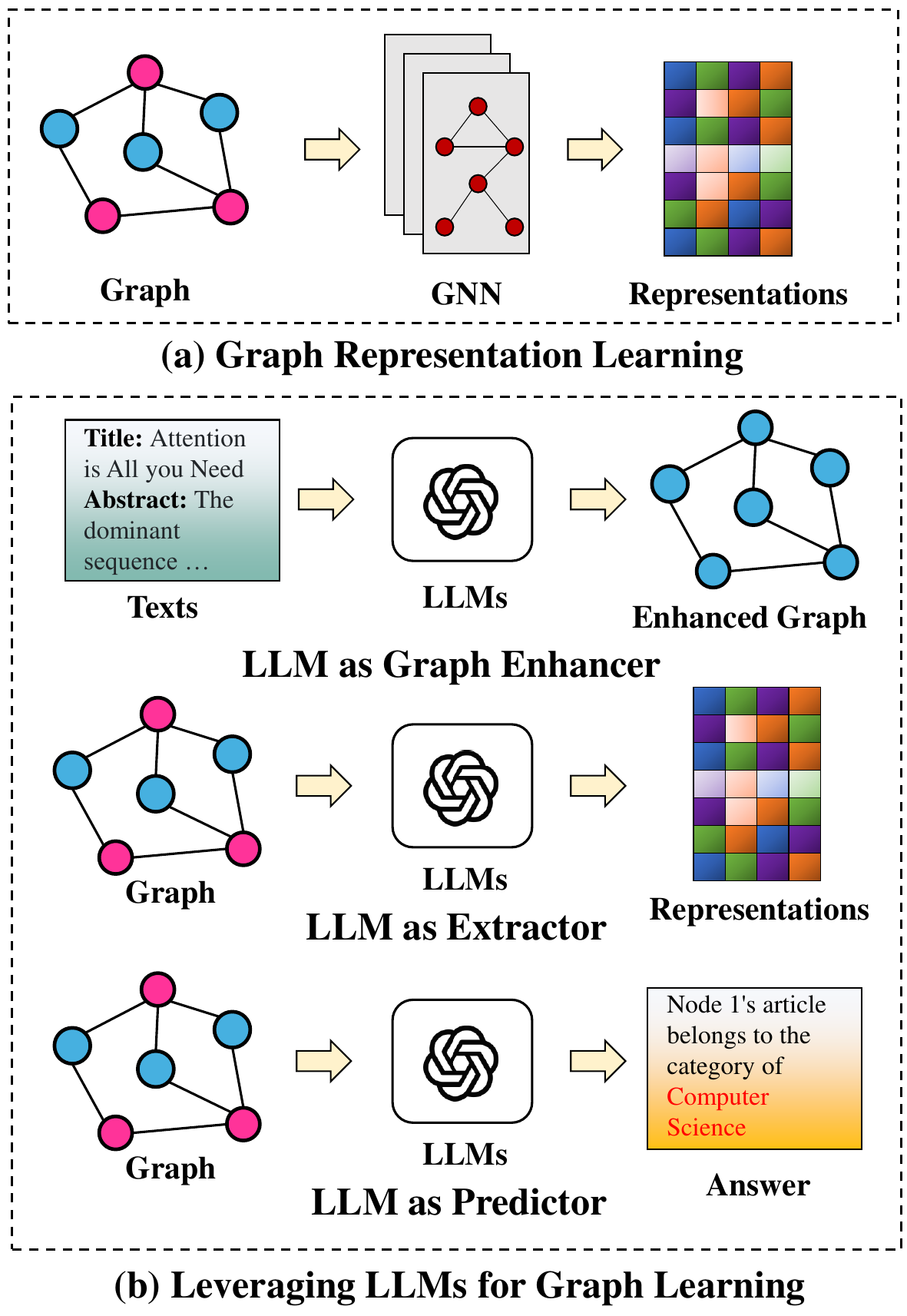}}
\caption{Comparison between traditional graph representation learning and emerging paradigms that leverage LLMs.
(a) Conventional graph learning relies on GNNs to encode structural information into node or graph-level representations.
(b) In LLM-based approaches, LLMs can enhance graph structures using auxiliary text (top), extract semantic representations directly from graph inputs (middle), or generate task-specific answers through reasoning over graph structures (bottom).}
\label{fig2}
\end{figure}

\subsection{Graph Notation Definition}

We denote a graph as $\mathcal{G} = (\mathcal{V}, \mathcal{E}, \mathbf{X}_V, \mathbf{X}_E)$, where $\mathcal{V}$ is the set of $n$ nodes, $\mathcal{E}$ is the set of edges, $\mathbf{X}_V \in \mathbb{R}^{|\mathcal{V}| \times d_v}$ is the node feature matrix, $\mathbf{X}_E \in \mathbb{R}^{|\mathcal{E}| \times d_e}$ is the edge feature matrix, and $\mathbf{A} \in \mathbb{R}^{n \times n}$ is the adjacency matrix, where $\mathbf{A}_{ij} = 1$ if $(v_i, v_j) \in \mathcal{E}$. The degree matrix is $\mathbf{D}$ with $\mathbf{D}_{ii} = \sum_j \mathbf{A}_{ij}$. The normalized graph Laplacian is defined as: $\mathbf{L} = \mathbf{I} - \mathbf{D}^{-1/2} \mathbf{A} \mathbf{D}^{-1/2}.$

The types of graph data involved in this work can be broadly categorized as follows.
Homogeneous graph refers to a graph in which all nodes and edges belong to the same type.
Heterogeneous graph refers to a graph where nodes and/or edges belong to multiple types, such as user/item nodes or different edge relations.
Attributed graph refers to a graph where nodes and/or edges are associated with real-valued features, \emph{i.e.}, $\mathbf{X}_V$ and/or $\mathbf{X}_E$ are non-empty.
Knowledge graph refers to a directed, multi-relational graph represented as triplets $(h, r, o)$, where $h \in \mathcal{V}$ is the head entity, $o \in \mathcal{V}$ is the object entity, and $r \in \mathcal{R}$ is a relation type.
Dynamic graph refers to a sequence of time-evolving graphs $\{\mathcal{G}_t\}_{t=1}^{T}$ in which the node set, edge set, or features may change over time.

\subsection{Graph Learning Methods}

Graph learning refers to a family of machine learning techniques designed specifically to analyze and learn from graph-structured data \cite{xia2021graph, 10.1145/3565028, khoshraftar2024survey}. Unlike data in Euclidean domains, graphs encode relationships between entities, and thus graph learning algorithms must leverage both node features and the graph topology. The goal is often to learn vector representations for nodes, edges, or entire subgraphs/graphs that capture their structural role and feature information, so that these embeddings can be used for downstream tasks, as shown in Figure~\ref{fig2}(a).

\textbf{Traditional Graph Learning Methods:} A wide range of traditional techniques have been developed to learn graph representations. Early methods such as spectral clustering \cite{ng2001spectral} and LINE \cite{tang2015line} learn embeddings by factorizing adjacency or similarity matrices, capturing global connectivity patterns. Random-walk-based approaches like DeepWalk \cite{perozzi2014deepwalk}, Node2Vec \cite{grover2016node2vec}, and metapath2vec \cite{dong2017metapath2vec} model local contexts through node sequences generated by truncated walks, offering better scalability and flexibility. In parallel, autoencoder-based models aim to reconstruct graph structures from latent representations. Notable examples include Graph Autoencoder (GAE) and Variational Graph Autoencoder (VGAE) \cite{kipf2016variational}, which adopt graph convolutional encoders with deterministic or probabilistic decoders. Further developments such as Adversarially Regularized Graph Autoencoders (ARGA) \cite{pan2018adversarially} and Graph Attention Autoencoders (GATE) \cite{salehi2019graph} introduce adversarial training and attention mechanisms to improve expressiveness and robustness. Despite their foundational role, traditional methods often fall short in handling the complexity of real-world graph data.

\textbf{Graph Neural Networks:} The advent of graph neural networks (GNNs) marked a significant change by enabling end-to-end learning directly on graph structures. Spectral GNNs, starting with Bruna \emph{et al.} \cite{bruna2014spectral}, were initially limited by high computational costs, which were mitigated by subsequent works using polynomial approximations \cite{defferrard2016convolutional}. The introduction of Graph Convolutional Networks (GCNs) by Kipf and Welling \cite{kipf2017semisupervised} popularized neighborhood aggregation in semi-supervised settings. Further developments such as GraphSAGE \cite{hamilton2017inductive}, GAT \cite{velivckovic2018graph}, SGC \cite{wu2019simplifying}, and GIN \cite{xu2019how} improved the expressiveness, scalability, and adaptability of GNNs.

More recently, two powerful approaches have emerged in graph learning: contrastive learning and Transformer-based architectures. Graph contrastive learning methods like DGI \cite{velickovic2019deep}, GraphCL \cite{you2020graph}, and HeCo \cite{wang2021self} focus on learning robust representations by maximizing consistency across multiple views of the same graph. Transformer-inspired models, such as GT \cite{shi2020masked} and GraphGPS \cite{rampavsek2022recipe}, introduce global self-attention and hybrid message-passing to model long-range dependencies and scale across graph types.

While graph learning techniques have shown remarkable success in modeling relational structures and topological dependencies, they often fall short in capturing high-level semantics and leveraging external unstructured data \cite{ren2024survey, yang2024state}. These limitations become more pronounced when dealing with real-world graphs that are frequently incomplete, imbalanced, heterogeneous across domains, or dynamically evolving over time. These challenges have motivated a growing interest in integrating graph models with powerful language-based representations, such as LLMs, which offer strong semantic reasoning capabilities and the flexibility to incorporate external knowledge into the graph learning pipeline.

\subsection{LLMs for Graph Learning}

LLMs are a class of deep neural networks pretrained on massive text corpora to capture the statistical and semantic structure of natural language \cite{naveed2023comprehensive, chang2024survey}. By learning distributed token representations and contextual relationships, LLMs achieve impressive performance across a broad spectrum of NLP tasks, including text generation \cite{li2024pre}, question answering \cite{singhal2025toward}, and machine translation \cite{moslem2023adaptive}.

The development of LLMs reflects a paradigm shift from task-specific pipelines to general-purpose language understanding. Early NLP relied on symbolic or statistical models (\emph{e.g.}, HMMs \cite{rabiner1989tutorial}, CRFs \cite{lafferty2001conditional}), which were constrained by domain specificity and limited context. The introduction of the Transformer architecture \cite{vaswani2017attention} enabled global attention and deep sequence modeling, which laid the foundation for modern LLMs. Subsequent milestones such as BERT \cite{devlin2019bert}, GPT-3 \cite{brown2020language}, and ChatGPT \cite{ouyang2022training} progressively advanced context modeling, few-shot reasoning, and conversational alignment. Recent systems like GPT-4 \cite{achiam2023gpt} and DeepSeek-MoE \cite{dai2024deepseekmoe} further scale capacity and introduce modular expert routing, enabling more efficient and multimodal learning.

Despite being originally designed for natural language, LLMs exhibit emergent capabilities—such as abstract reasoning, flexible generalization, and heterogeneous input integration—that extend far beyond textual tasks. As shown in Figure~\ref{fig2}(b), LLMs can complement graph learning by generating structural hypotheses \cite{zhang2025darg}, enriching semantic content \cite{brasoveanu2023framing}, and aligning cross-domain knowledge \cite{liu2023evaluating}. Viewed through the lens of graph learning, these abilities make LLMs well-suited for enhancing data quality and robustness in graph-based applications:

\textbf{Semantic reasoning:} By leveraging rich contextual understanding, LLMs can infer latent or missing information from incomplete inputs. In the graph domain, this enables the imputation of missing node or edge attributes, the prediction of potential links, and the interpretation of sparsely connected substructures. Such capabilities directly address the challenge of \textbf{incompleteness} in real-world graphs, where data sparsity or noise is common \cite{brasoveanu2023framing}.

\textbf{Cross-domain and cross-modal alignment:} LLMs are trained on diverse sources of knowledge and can serve as translators between incompatible domains or feature spaces. This makes them particularly useful in \textbf{cross-domain heterogeneous} graphs \cite{liu2023evaluating}—for example, aligning product graphs across e-commerce platforms that differ in attribute schemas, category taxonomies, and language conventions. LLMs can bridge such gaps by reasoning over textual descriptions and contextual cues to identify semantically equivalent items across domains.

\textbf{Generative augmentation:} As generative models, LLMs can synthesize new graph components—such as nodes, edges, or subgraphs—conditioned on textual or structural prompts. This ability can be used to create balanced training samples for underrepresented classes, thereby mitigating \textbf{data imbalance} \cite{yu2025leveraging}, or to simulate hypothetical future graph states for handling \textbf{dynamic instability} \cite{zhang2025darg}. 

In the following sections, we examine each of the four fundamental data-centric challenges in turn, analyzing how LLMs have been used to mitigate them, and what open questions remain at this intersection.

\section{Current Techniques for Graph Challenges} \label{s3}
In this section, we discuss each of the four core challenges in detail. For each challenge, we first formalize the problem. We then review representative traditional and LLM-enhanced solutions to provide context.

\subsection{Incompleteness in Graphs}
\begin{figure}[t]
\centerline{\includegraphics[width=1\linewidth]{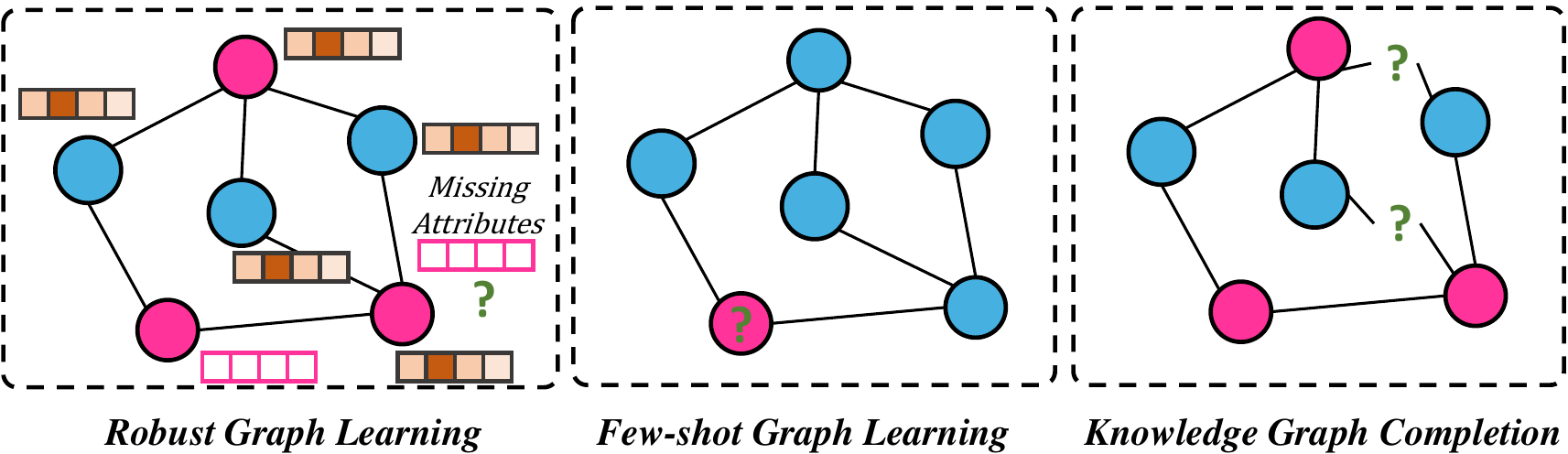}}
\caption{Illustrations of three representative tasks addressing graph incompleteness.
Robust Graph Learning (left) handles noisy or missing attributes to ensure stable performance.
Few-shot Graph Learning (middle) aims to generalize from limited labeled nodes or subgraphs.
Knowledge Graph Completion (right) focuses on inferring missing links between entities based on observed relationships.}
\label{fig4}
\end{figure}

In the real world, graph data is often incomplete due to missing node attributes, edges or labels,  which significantly impacts the accuracy and generalization ability of graph models. By leveraging LLMs for graph data completion and the inference of missing information, the performance of graph learning models can be significantly improved, even in scenarios with incomplete or missing information. We categorize current approaches to incomplete graphs into three major directions (as shown in Figure \ref{fig4}): (1) Robust Graph Learning, which focuses on making models resilient to missing data; (2) Few-Shot Graph Learning, which addresses scenarios with extremely limited labeled data or structure by transferring knowledge; and (3) Knowledge Graph Completion, which is a prominent sub-area dealing with inferring missing facts in large knowledge graphs. The relevant references and categorization are presented in Table \ref{tab:incompleteness_methods}.

\begin{table*}[htbp]\scriptsize
  \centering
  \caption{LLM-based methods for handling \textbf{incompleteness} in graphs, grouped by domain and incomplete type, with representative methods, datasets, metrics, and downstream tasks.}
  \renewcommand{\arraystretch}{2}
  \setlength{\tabcolsep}{3.5pt}
  \begin{adjustbox}{max width = 1.0\linewidth}
  \begin{tabular}{c P{0.1\linewidth} P{0.1\linewidth} P{0.3\linewidth} P{0.15\linewidth} P{0.15\linewidth}}
    \toprule
    \textbf{Domain} & \textbf{Incompleteness} & \textbf{Method} & \textbf{Typical Datasets} & \textbf{Common Metrics} & \textbf{Downstream Tasks} \\
    \midrule
    
    \multirow{10}{*}{\makecell[c]{Robust \\Graph \\Learning}}
    & \multirow{3}{*}{Node}
      & \textbf{LLM4NG}~\cite{yu2025leveraging} & Cora~\cite{mccallum2000automating}, PubMed~\cite{sen2008collective}, ogbn-arxiv \cite{hu2020open} & Accuracy& Node Classification \\
    &   & \textbf{LLM-TAG}~\cite{sun2023large}& Cora \cite{mccallum2000automating}, Citeseer \cite{giles1998citeseer}, PubMed \cite{sen2008collective}, Arxiv-2023 \cite{he2024harnessing} & Accuracy & Node Classification \\
    &   & \textbf{SPLLM}~\cite{fang2025spatiotemporal} &PeMS03 \cite{zhang2021traffic}, PeMS04 \cite{zhang2021traffic}, PeMS07 \cite{zhang2021traffic} &MAE, RMSE, MAPE & Spatiotemporal Forecasting\\
    \cmidrule(lr){2-6}
    & Label
      & \textbf{LLMGNN}~\cite{chen2023label} &Cora \cite{mccallum2000automating}, Citeseer \cite{giles1998citeseer}, PubMed \cite{sen2008collective}, ogbn-arxiv \cite{hu2020open}, ogbn-products \cite{hu2020open}, Wikics \cite{mernyei2020wiki} &Accuracy & Node Classification\\
    \cmidrule(lr){2-6}
    & \multirow{4}{*}{Mixed}
      & \textbf{GraphLLM}~\cite{chai2023graphllm} & Synthetic Data \cite{wang2023can} &Exact Match Accuracy & Graph Reasoning\\
    &   & \textbf{PROLINK}~\cite{wang2024llm} & FB15k237 \cite{toutanova2015observed}, Wikidata68K \cite{GeseseSA22}, NELL-995 \cite{XiongHW17}&MRR, Hits@N&Knowledge Graph Completion \\
    &   & \textbf{UnIMP}~\cite{wang2025llm} &BG \cite{asuncion2007uci}, ZO \cite{asuncion2007uci}, PK \cite{asuncion2007uci}, BK \cite{asuncion2007uci}, CS \cite{asuncion2007uci}, ST \cite{asuncion2007uci}, PW \cite{asuncion2007uci}, BY \cite{asuncion2007uci}, RR \cite{asuncion2007uci}, WM \cite{asuncion2007uci} &RMSE, MAE & Data Imputation\\

    \midrule

    \multirow{9}{*}{\makecell[c]{Few-Shot \\Graph \\Learning}}
    & \multirow{2}{*}{Structure}
      & \textbf{LinkGPT}~\cite{he2024linkgpt} &AmazonSports \cite{mcauley2015image}, Amazon-Clothing \cite{mcauley2015image}, MAG-Geology \cite{sinha2015overview}, MAG-Math \cite{sinha2015overview} &MRR, Hits@N & Link Prediction\\
    &   & \textbf{AnomalyLLM}~\cite{liu2024anomalyllm} & UCI Messages \cite{opsahl2009clustering}, Blogcatalog \cite{tang2009relational}, T-Finance \cite{tang2022rethinking}, T-Social \cite{tang2022rethinking}& AUC &Anomaly Detection \\
    \cmidrule(lr){2-6}
    & \multirow{4}{*}{Mixed}
      & \textbf{LLMDGCN}~\cite{li2024llm} & Cora \cite{mccallum2000automating}, Citeseer \cite{giles1998citeseer}, PubMed \cite{sen2008collective}, Religion \cite{rosenthal2016social} &Accuracy & Node Classification\\
    &   & \textbf{HeGTa}~\cite{jin2024hegta} &IM-TQA \cite{zheng2023tqa}, WCC \cite{ghasemi2018tabvec}, HiTab \cite{cheng2021hitab}, WTQ \cite{pasupat2015compositional}, TabFact \cite{chen2019tabfact} &Macro-F1, Accuracy & Table Understanding\\
    &   & \textbf{FlexKBQA}~\cite{li2024flexkbqa} &GrailQA \cite{gu2021beyond}, WebQSP \cite{yih2016value}, KQA Pro \cite{cao2022kqa} &Exact Match, F1, Accuracy & Knowledge Graph Question Answering\\
    &   & \textbf{KGQG}~\cite{zhao2024zero} & WebQuestions \cite{kumar2019difficulty}, PathQuestions \cite{zhou2018interpretable} &BLEU-4, ROUGE-L, Hits@N & Knowledge Graph Question Answering\\

    \midrule

    \multirow{22}{*}{\makecell[c]{Knowledge \\Graph \\Completion}}
    & \multirow{3}{*}{Node}
      & \textbf{LLM-KGC}~\cite{sehwag2024context} &ILPC \cite{galkin2022open} &MRR, Hits@N &Knowledge Graph Completion \\
    &   & \textbf{GS-KGC}~\cite{yang2025gs} &WN18RR \cite{dettmers2018convolutional}, FB15k-237 \cite{toutanova2015observed}, FB15k-237N \cite{lv2022pre}, ICEWS14 \cite{garcíadurán2018learningsequenceencoderstemporal}, ICEWS05-15 \cite{li2021temporal}& Hits@N &Knowledge Graph Completion \\
    &   & \textbf{GLTW}~\cite{luo2025gltw} &FB15k-237 \cite{toutanova2015observed}, WN18RR \cite{dettmers2018convolutional}, Wikidata5M \cite{vrandevcic2014wikidata} & MRR, Hits@N & Link Prediction\\
    \cmidrule(lr){2-6}
    & Label
      & \textbf{KGs-LLM}~\cite{carta2023iterative} &Wikipedia \cite{carta2023iterative} &F1, Precision, Recall & Knowledge Graph Generation\\
    \cmidrule(lr){2-6}
    & \multirow{12}{*}{Mixed}
      & \textbf{FSKG}~\cite{brasoveanu2023framing} &WN18RR \cite{dettmers2018convolutional}, FB15k-237 \cite{toutanova2015observed} &MRR, Hits@N & Knowledge Graph Completion\\
    &   & \textbf{KGLLM}~\cite{yao2023exploring} &WN11 \cite{socher2013reasoning}, FB13 \cite{socher2013reasoning}, WN18RR \cite{dettmers2018convolutional}, YAGO3-10 \cite{dettmers2018convolutional} &Accuracy, MRR, Hits@N&Link Prediction, Knowledge Graph Completion \\
    &   & \textbf{KICGPT}~\cite{wei2023kicgpt} &FB15k-237 \cite{toutanova2015observed}, WN18RR \cite{dettmers2018convolutional} &MRR, Hits@N &Link Prediction \\
    &   & \textbf{RL-LLM}~\cite{chen2023knowledge} &Electronics, Instacart \cite{chen2023knowledge} & Precision, Recall, Accuracy&Knowledge Graph Completion \\
    &   & \textbf{GoG}~\cite{xu2024generate} &Synthetic Data \cite{xu2024generate} & Hits@N& Knowledge Graph Question Answering\\
    &   & \textbf{KoPA}~\cite{zhang2024making} & UMLS \cite{yao2019kg}, CoDeX-S \cite{lv2022pre}, FB15K-237N \cite{lv2022pre} &F1, Precision, Recall, Accuracy  & Knowledge Graph Completion\\
    &   & \textbf{LLMKG}~\cite{iga2024assessing} &Templates Easy \cite{iga2024assessing}, Templates Hard \cite{iga2024assessing} & Strict Metrics, Flexible Metrics&Knowledge Graph Completion \\
    &   & \textbf{DIFT}~\cite{liu2024finetuning} & WN18RR \cite{dettmers2018convolutional}, FB15k-237 \cite{toutanova2015observed} &MRR, Hits@N & Link Prediction, Knowledge Graph Completion \\
    &   & \textbf{CP-KGC}~\cite{yang2024enhancing} &WN18RR \cite{dettmers2018convolutional}, FB15k-237 \cite{toutanova2015observed}, UMLS \cite{yao2019kg} & MRR, Hits@N &Knowledge Graph Completion \\
    &   & \textbf{MuKDC}~\cite{li2024llmb} & NELL \cite{mitchell2018never}, Wiki \cite{vrandevcic2014wikidata} & MRR, Hits@N &Knowledge Graph Completion  \\

    \bottomrule
  \end{tabular}
  \end{adjustbox}
  \label{tab:incompleteness_methods}
\end{table*}

\subsubsection{Robust Graph Learning}
Robust graph learning methods aim to maintain model performance even when the input graph is noisy or partially observed. Traditional GNNs often degrade in accuracy if key attributes are missing or if the graph is sparsely connected, because the iterative message passing has less information to propagate. Therefore, researchers have proposed specialized techniques to make graph learning robust to incompleteness.

\textbf{(1)Traditional Methods for Robust Graph Learning}
Early efforts in robust graph learning focused on imputing missing values using global interpolation techniques under low-rank assumptions. For example, matrix completion methods \cite{csimcsek2008navigating, cai2010singular, huang2019graph, chen2019attributed} estimate missing attributes based on the global structure of the data. GraphRNA \cite{huang2019graph} introduces Attribute Random Walk (AttriWalk), modeling attributes as a bipartite graph to capture attribute-structure interactions, thereby enhancing local adaptability and robustness in heterogeneous environments. ARWMF \cite{chen2019attributed} combines random walks with matrix factorization and mutual information to improve unsupervised node embedding under noisy conditions.

With the rise of deep learning, more sophisticated generative models were developed to overcome the linear assumptions of early methods \cite{chen2022learning, yoo2022accurate, tu2022initializing, morales2022simultaneous, peng2024multi, xia2024attribute, li2024csat, li2024scae, li2025topology, chen2025temporal}. SAT \cite{chen2022learning} enhances robustness by decoupling structure and attribute signals via shared latent spaces and distribution alignment. SVGA \cite{yoo2022accurate} imposes strong probabilistic regularization on latent variables to prevent overfitting when features are sparse or unreliable. Models like ITR \cite{tu2022initializing} and VISL \cite{morales2022simultaneous} progressively refine latent variables or inter-variable structures using topological cues, improving resilience against structural noise. More recently, methods like MATE \cite{peng2024multi} and AIAE \cite{xia2024attribute} exploit multi-view and multi-scale generation strategies to stabilize representation learning under incomplete or corrupted input. CSAT \cite{li2024csat} incorporates contrastive learning and Transformers to detect communities under noisy or weak supervision, further enhancing robustness through auxiliary signals.

Another line of work improves model robustness through dynamic inference mechanisms that rely on structural priors like homophily and community consistency \cite{rossi2022unreasonable, um2023confidence, li2025attrireboost}. For instance, FP \cite{rossi2022unreasonable} integrates Dirichlet energy minimization with graph diffusion to achieve stable feature recovery. PCFI \cite{um2023confidence} assigns confidence scores to feature channels, allowing uncertainty-aware propagation. ARB \cite{li2025attrireboost} addresses the cold-start problem by introducing virtual edges and redefined boundary conditions to improve propagation in sparse or poorly connected graphs. A notable advantage of such propagation-based methods is their parameter-free nature, making them highly adaptable and easily integrated with LLM-based pipelines for downstream reasoning tasks. To tackle the cold-start problem in node representation learning, Cold Brew \cite{zheng2021cold} introduced a feature contribution ratio metric to guide teacher-student distillation, uniquely encoding topological sparsity as a dynamic temperature coefficient. CTAug \cite{wu2024graphlmr} employs subgraph-aware contrastive learning to preserve dense subgraph priors, significantly enhancing representation learning in dense graphs.

These methods significantly enhance the predictive capabilities of graph data in scenarios with missing attributes by jointly modeling, employing variational inference, and optimizing multi-source information. Nevertheless, three challenges remain: (1) coupling optimization of completion and prediction dramatically increases training complexity; (2) latent-space completion methods are sensitive to prior assumptions; (3) efficient incremental mechanisms are lacking for dynamic attribute completion.

\textbf{(2) LLM-enhanced methods}
Recently, researchers have begun exploring the deep integration of semantic reasoning and incomplete graph learning. Methods based on LLMs \cite{sun2023large, yu2025leveraging, fang2025spatiotemporal, chai2023graphllm, wang2024llm, chen2023label} pioneer a new paradigm that combines semantic completion and structural refinement, leveraging the strong generative and reasoning capabilities of LLMs.

\begin{figure}[t]
\centerline{\includegraphics[width=0.9\linewidth]{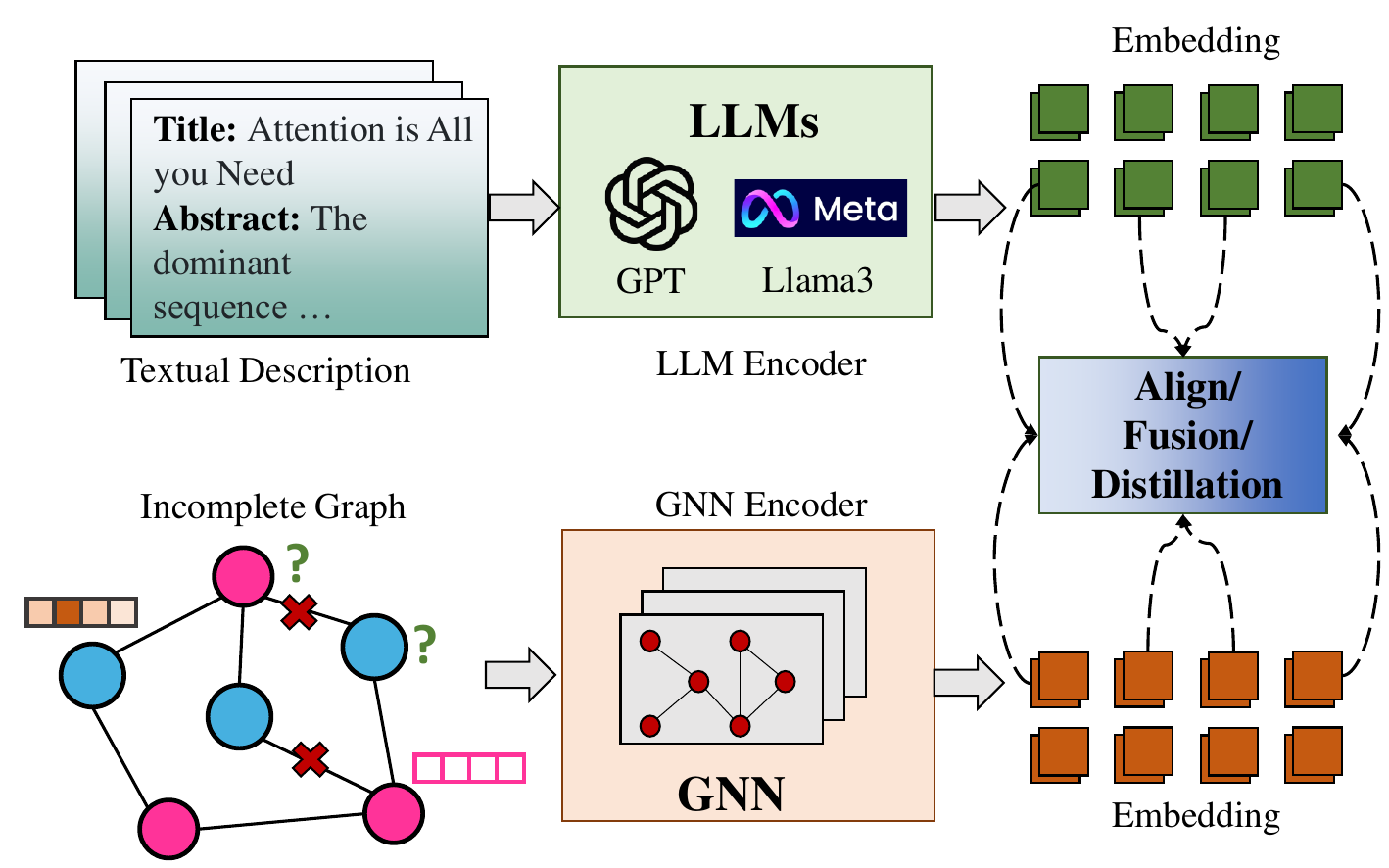}}
\caption{The framework addresses incomplete graph structures by integrating structural signals from GNNs and semantic priors from LLMs. The ``Align / Fusion / Distillation” module enables the model to compensate for missing graph information by leveraging textual context, resulting in enriched and unified node attribute/edge/label representations.}
\label{fig3}
\end{figure}

Figure~\ref{fig3} illustrates how LLMs can extract domain-specific knowledge and use it to compensate for missing nodes, edges, or attributes. For example, Sun \emph{et al.} \cite{sun2023large} demonstrate that LLMs can infer missing nodes and edges by recognizing semantic similarities between entities. In this way, the model does not just ``guess” a connection statistically, but proposes links that are meaningful in the real world. Building on this idea, LLM4NG \cite{yu2025leveraging} generates new, contextually relevant nodes, which is especially helpful in few-shot settings where the graph is very sparse. Both methods show that LLMs can enrich incomplete graphs more effectively than traditional imputation. Meanwhile, SPLLM \cite{fang2025spatiotemporal} focuses on traffic sensor networks, where missing values break both spatial and temporal dependencies. By combining spatiotemporal GCNs with LLM fine-tuning, and incorporating external knowledge such as road maps or weather data, it achieves more reliable predictions. Similarly, UnIMP \cite{wang2025llm} works on tabular data by treating it as a hypergraph and using LLMs to fill in missing values, even when the data is semantically or structurally heterogeneous. 

To further close the semantic gap introduced by graph incompleteness, several works propose end-to-end frameworks combining LLMs and graph models. For instance, GraphLLM \cite{chai2023graphllm} introduces a unified architecture where a graph encoder maps node features into the LLM’s semantic space, enabling it to reason over incomplete or noisy node attributes using attention-gated fusion of structure and context. This helps, for example, in citation networks or recommendation systems where node metadata may be sparse or outdated. PROLINK \cite{wang2024llm} enhances inductive reasoning on low-resource knowledge graphs by generating structural prompts and using LLMs to fill in gaps from limited graph and text data — particularly valuable in biomedical or emerging domains with sparse curation. Lastly, LLMGNN \cite{chen2023label} proposes a semi-supervised approach where LLMs annotate a small subset of nodes, helping GNNs generalize from minimal labeled data — addressing the classic label incompleteness challenge in large-scale graphs.

\subsubsection{Few-shot Graph Learning}
Few-shot learning (FSL) deals with scenarios where very few labeled examples are available for training, or where a graph has only a handful of nodes/edges in a particular class or subset of interest. This is related to incompleteness in that labels can be viewed as a type of missing information.

\textbf{(1) Traditional Methods for Few-Shot Learning}

FSL in graph domains aims to generalize effectively from limited labeled samples, addressing not only label scarcity but also structural issues such as missing links and long-tail relations \cite{xiong2018one, yao2020graph, zhang2020few, sheng2020adaptive, qian2021distilling}. Early approaches primarily focused on explicitly modeling local graph structures and relational semantics under low-resource settings. For instance, Gmatching \cite{xiong2018one} pioneered the integration of single-hop neighborhood information with relation prototype matching, providing a foundation for rare relation prediction. Building on this, GFL \cite{yao2020graph} established a transferable metric space by projecting auxiliary knowledge into the target domain, improving cross-graph adaptability.

Subsequent works moved toward more sophisticated meta-learning strategies. FSRL \cite{zhang2020few} introduced a heterogeneous relation meta-learning framework that decouples relation semantics through a reference representation generator, enabling better handling of heterogeneous relation types. Complementing this semantic focus, FAAN \cite{sheng2020adaptive} designed an adaptive encoder–aggregator mechanism to dynamically weight entity-reference contributions, thereby producing fine-grained semantic representations that significantly enhance few-shot knowledge graph completion. Extending FSL techniques to real-world applications, MetaHG \cite{qian2021distilling} demonstrated the utility of meta-learning in heterogeneous environments by detecting drug trafficking on Instagram.

Despite these advances, traditional FSL methods remain constrained by hand-crafted architecture designs, often lacking efficiency when dealing with large, complex graph structures and exhibiting limited capability for structured and interpretable reasoning.

\textbf{(2) Few-Shot Learning in LLMs}
Few-shot learning has emerged as a natural strength of LLMs, enabling them to handle graph-related tasks with limited labeled data by leveraging rich prior knowledge from pretraining. Recent work demonstrates their effectiveness in diverse scenarios such as graph classification \cite{li2024llm}, link prediction \cite{he2024linkgpt}, anomaly detection \cite{liu2024anomalyllm}, table understanding \cite{jin2024hegta}, and knowledge graph question answering (KGQA) \cite{li2024flexkbqa,zhao2024zero}.
For instance, LLMDGCN \cite{li2024llm} incorporates degree-aware prompt tuning with graph-encoded positional embeddings, enabling both node classification and edge recovery in low-label settings. In link prediction, LinkGPT \cite{he2024linkgpt} combines instruction tuning with retrieval-based re-ranking to enhance reasoning under sparse supervision. AnomalyLLM \cite{liu2024anomalyllm} integrates dynamic-aware encoding and prototype-based edge reprogramming for improved detection in dynamic graphs. Beyond pure graph tasks, HeGTa \cite{jin2024hegta} aligns table semantics with LLM knowledge via heterogeneous graph-enhanced soft prompts, while FlexKBQA \cite{li2024flexkbqa} and a zero-shot KGQA approach \cite{zhao2024zero} reformulate KB queries into natural language to generate synthetic training data or reduce reliance on full supervision. Collectively, these results indicate that LLMs can generalize to novel graph tasks with minimal examples, acting as strong priors that bridge missing or noisy information.

In summary, while few-shot graph learning remains challenging, LLMs act as few-shot learners by nature – they can leverage their pretrained knowledge to make sense of novel tasks with limited data. Results show that even with minimal graph information, an LLM can propose connections or classifications that align with human knowledge, essentially providing a powerful prior to the graph model.

\subsubsection{Knowledge Graph Completion}
Incompleteness is inherent in knowledge graphs: no knowledge base is ever complete. The development of knowledge graph completion (KGC) revolves around the core challenge of structure-semantics fusion, forming a progressive innovation trajectory from fundamental approach exploration to multimodal collaboration. LLMs, with their strong language understanding and reasoning capabilities, demonstrate great potential in addressing long-tail challenges, reducing annotation burdens, and handling incomplete graph data \cite{carta2023iterative, brasoveanu2023framing, yao2023exploring,  chen2023knowledge, xu2024generate, sehwag2024context, zhang2024making, li2024llmb, iga2024assessing, liu2024finetuning, yang2024enhancing, yang2025gs, wei2023kicgpt}.

Early explorations of LLM-driven KGC primarily focused on converting triples into natural language sequences to verify whether generative LLMs could address incompleteness effectively. FSKG \cite{brasoveanu2023framing} exemplifies this trend by introducing a generation–refinement pipeline that mitigates long-tail sparsity through staged generation. Building on this concept, MuKDC incorporated multi-stage knowledge distillation to generate coherent supplementary triples, further improving the coverage of long-tail relations. Extending beyond generation quality, KGs-LLM \cite{carta2023iterative} demonstrated that zero-shot prompting, even without external knowledge, can iteratively extract graph components, thereby reducing annotation costs and scaling efficiently to domain-specific datasets.
While these methods validated the feasibility of text-based KGC, KGLLM \cite{yao2023exploring} further showed that such reformulation enables lightweight fine-tuning of smaller LLMs like LLaMA-7B, achieving competitive triple classification and relation prediction. Addressing the persistent challenge of long-tail entities, KICGPT \cite{wei2023kicgpt} integrated a triple-based retriever with contextual prompting, directly encoding structural cues into LLM inputs. This approach avoids additional training while yielding strong few-shot performance, thus highlighting the efficiency of retrieval-augmented prompting in KGC.

To mitigate the structural information loss inherent in purely text-based approaches, researchers have developed prompt encoding strategies that embed richer graph context. RL-LLM \cite{chen2023knowledge} applied few-shot learning with multi-prompt optimization for relation prediction, showing that even minimal labeled data can yield competitive results in e-commerce KGs. Expanding the reasoning process, GoG \cite{xu2024generate} proposed a ``think–search–generate” pipeline, enabling the synthesis of new triples without additional training, which strengthens reasoning over incomplete graphs. Complementing this, LLM-KGC \cite{sehwag2024context} fused ontology and graph structure directly into prompts, aligning topological and semantic information to improve inductive reasoning.
Further enhancing structure-aware inference, KoPA \cite{zhang2024making} introduced a knowledge prefix adapter to inject structural signals during LLM reasoning, thereby improving logical consistency in KGC predictions. These works collectively highlight a shift from simple text reformulation toward structurally enriched prompting, aiming to bridge the gap between LLM semantic priors and graph topology.

Building upon these prompting and structural integration advances, recent research has optimized LLM–graph fusion for few-shot KGC, entity alignment, and robustness in heterogeneous settings. MuKDC \cite{li2024llmb} extended its earlier generation framework by incorporating multimodal knowledge and consistency evaluation, boosting performance on long-tail few-shot tasks. In parallel, LLMKG \cite{iga2024assessing} demonstrated that carefully engineered prompts allow both Mixtral-8x7b and GPT-4 to perform competitively in zero- and one-shot KGC across diverse metrics. Addressing entity alignment, DIFT \cite{liu2024finetuning} combined lightweight models with discriminative prompts to prevent alignment drift while improving overall KGC accuracy. In handling semantic ambiguity, CP-KGC \cite{yang2024enhancing} leveraged contextually adaptive prompts to disambiguate polysemous entities, particularly in quantized LLMs. Moving toward deeper structural integration, GLTW \cite{luo2025gltw} fused Graph Transformers with LLMs to jointly capture local and global patterns, while GS-KGC \cite{yang2025gs} employed subgraph-based question answering with negative sampling to enhance missing triple identification. Together, these frameworks signal a transition from isolated prompt design to holistic, multimodal, and structure-aware LLM–graph systems.

Overall, LLMs bring in a semantic understanding of the KG elements, which is valuable because KGs often have textual labels for entities/relations that carry meaning. LLMs leverage this to make more informed predictions, especially when data is sparse or the pattern is not purely structural but semantic.

\subsubsection{Evaluation for Handling Incompleteness in Graphs}

We summarize the existing evaluation pipeline of LLM-based methods for incomplete graph learning, covering benchmark datasets, evaluation metrics, and downstream tasks (see Table~\ref{tab:incompleteness_methods}).

Across the reviewed methods, commonly used benchmark datasets include Cora \cite{mccallum2000automating}, Citeseer~\cite{giles1998citeseer}, PubMed \cite{sen2008collective}, and ogbn-arxiv \cite{hu2020open} for node classification tasks. These are citation networks where each node is a document with associated textual attributes. For large-scale knowledge graph tasks, datasets such as FB15k-237 \cite{toutanova2015observed}, WN18RR \cite{dettmers2018convolutional}, Wikidata5M \cite{vrandevcic2014wikidata}, NELL \cite{mitchell2018never}, and UMLS \cite{yao2019kg} are widely adopted for link prediction and knowledge graph completion. Additionally, domain-specific or task-specific datasets are used to evaluate generalizability, including Amazon-Clothing and Amazon-Sports \cite{mcauley2015image} for recommendation and link prediction, PeMS03/04/07 \cite{zhang2021traffic} for spatiotemporal forecasting, TQA/WTQ/TabFact \cite{zheng2023tqa, pasupat2015compositional, chen2019tabfact} for table understanding, and GrailQA/WebQSP/KQA Pro \cite{gu2021beyond, yih2016value, cao2022kqa} for knowledge graph question answering. Some works also utilize synthetic graphs \cite{wang2023can, xu2024generate} to test reasoning capabilities in controlled environments.

A variety of evaluation metrics are used depending on the task. For classification, Accuracy, Macro-F1, and Micro-F1 are most common. Knowledge graph-related tasks report Mean Reciprocal Rank (MRR) and Hits@N as standard ranking metrics. Spatiotemporal forecasting relies on MAE, RMSE, and MAPE, while generative tasks such as question answering are evaluated using Exact Match, BLEU-4, and ROUGE-L.

The surveyed models address a wide range of downstream tasks, including node classification, link prediction, knowledge graph completion, anomaly detection, graph reasoning, question answering over knowledge graphs, and spatiotemporal forecasting. This diversity highlights the versatility of LLMs in handling incomplete graph data across domains and tasks.

\subsubsection{Summary of Incompleteness}

Across robust learning, few-shot learning, and KGC, the common thread is that LLMs serve as knowledge-infusers and intelligent guessers for what is missing in the graph. They help create a more complete picture by either filling in data directly or guiding the graph model on where to look. Empirically, studies report that incorporating LLM-generated features or suggestions leads to substantial gains in tasks like node classification with missing features, link prediction with sparse edges, and KG completion benchmarks. As LLMs continue to improve, especially in domain-specific knowledge, we expect their role in handling graph incompleteness to grow.

\subsection{Imbalance in Graphs}
\begin{figure}[htbp]
\centerline{\includegraphics[width=0.90\linewidth]{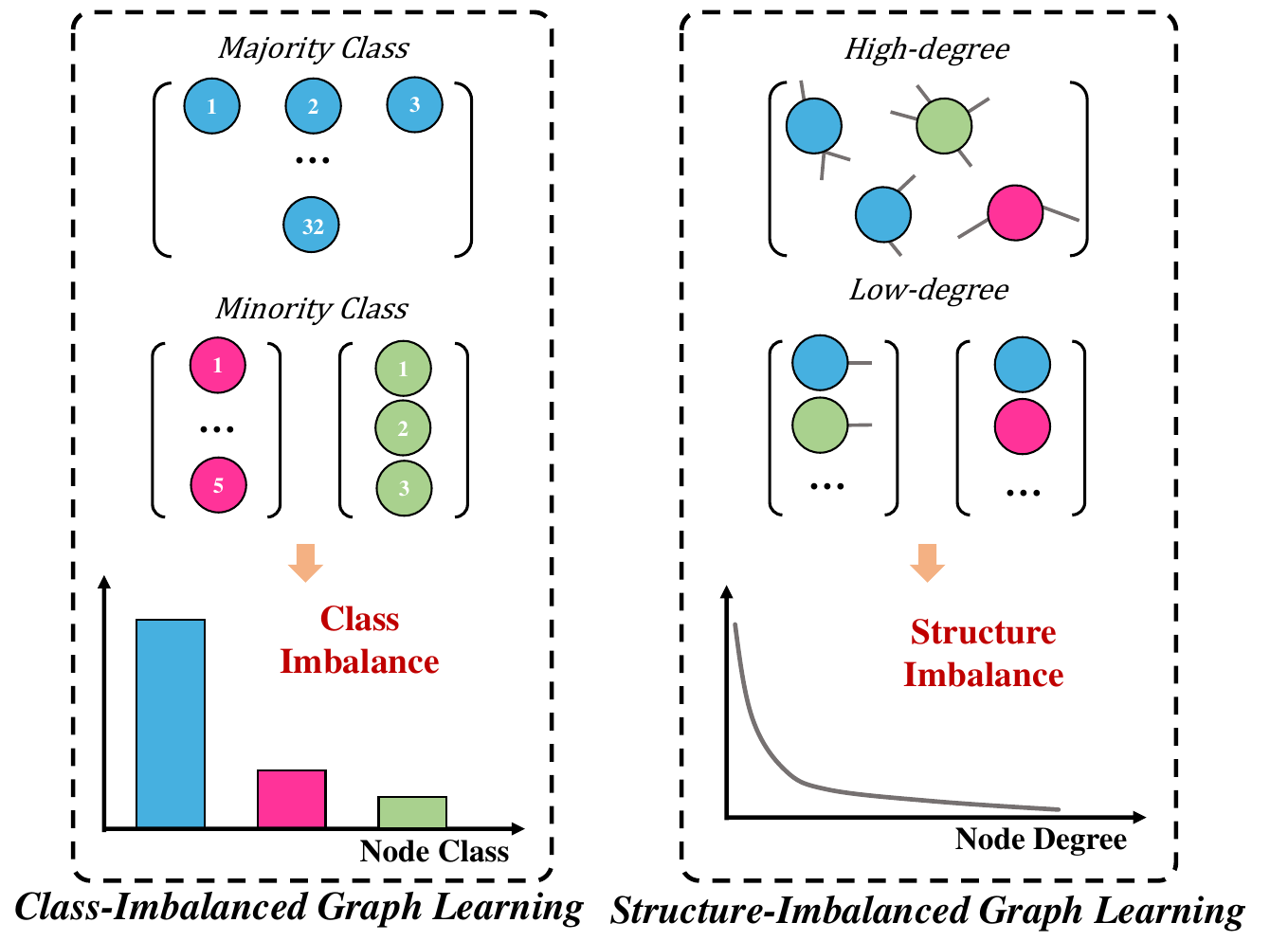}}
\caption{Illustration of three representative tasks addressing  graph imbalance. Class-imbalanced graph learning (left) refers to scenarios where certain node classes (minority classes) are underrepresented compared to others.
Structure-imbalanced graph learning (right) highlights disparities in node connectivity, where high-degree nodes dominate information flow while low-degree nodes are underexposed.}
\label{fig2-1}
\end{figure}

Graph data often exhibits imbalance that can severely affect the performance, particularly in tasks such as node classification (as shown in Figure \ref{fig2-1}). This imbalance typically complicates the learning process as models tend to be biased toward the majority class or low-degree nodes, resulting in poor generalization for underrepresented elements. To address this, techniques such as resampling, re-weighting, and graph-based regularization have been proposed to alleviate the effects of data imbalance. However, these approaches often require domain-specific adjustments and may not always yield optimal results.

Recent advances in LLMs offer promising solutions to the challenges posed by imbalanced graph data. LLMs, with their ability to process and generate semantically rich representations, can be leveraged to enrich the graph's feature space, enabling more nuanced and balanced learning. 

This section reviews the application of traditional methods and LLMs for learning from imbalanced graph data. The literature is categorized into two key research areas: (1) Class Imbalance and (2) Structure Imbalance. The relevant references and categorization are presented in Table \ref{tab:imbalance_methods}.

\begin{table*}[htbp]\scriptsize
  \centering
  \caption{LLM-based methods for handling \textbf{imbalance} in graphs, grouped by domain and task, with representative methods, datasets, metrics, and downstream tasks.}
  \renewcommand{\arraystretch}{2}
  \setlength{\tabcolsep}{3.5pt}
  \begin{adjustbox}{max width = 1.0\linewidth}
  \begin{tabular}{c P{0.1\linewidth} P{0.15\linewidth} P{0.25\linewidth} P{0.15\linewidth} P{0.15\linewidth}}
    \toprule
    \textbf{Domain} & \textbf{Tasks} & \textbf{Method} &\textbf{Typical Datasets} &\textbf{Common Metrics} & \textbf{Downstream Tasks} \\
    \midrule
    
    \multirow{28}{*}{\makecell[c]{Class\\ Imbalance\\ Graph \\Learning}}
    & \multirow{17}{*}{\makecell[c]{Node \\Classification}}
      & \textbf{LLM4NG}~\cite{yu2025leveraging} & Cora \cite{mccallum2000automating}, PubMed \cite{sen2008collective}, ogbn-arxiv \cite{hu2020open} & Accuracy& Few-shot Node Classification \\
    & & \textbf{LLM-GNN}~\cite{chen2023label} & Cora \cite{mccallum2000automating}, Citeseer~\cite{giles1998citeseer}, PubMed \cite{sen2008collective}, ogbn-arxiv \cite{hu2020open}, ogbn-products \cite{hu2020open}, Wikics \cite{mernyei2020wiki} &Accuracy & Label‐free Node Classification\\
    & & \textbf{G2P2}~\cite{wen2023augmenting} & Cora \cite{mccallum2000automating}, Art \cite{ni2019justifying}, Industrial \cite{ni2019justifying} and Music Instruments \cite{ni2019justifying} & Accuracy, Macro-F1 & Zero- and Few-shot Low-resource Text Classification  \\
    & & \textbf{LA-TAG}~\cite{wang2024large} & Cora \cite{mccallum2000automating}, PubMed \cite{sen2008collective}, Photo \cite{yan2023comprehensive}, Computer \cite{yan2023comprehensive}, and Children \cite{yan2023comprehensive} & Accuracy, Macro-F1 & Zero- and Few-shot Low-resource Text Classification \\
    & & \textbf{GSS-Net}~\cite{zhang2024fine} & Magazine Subscriptions \cite{ni2019justifying}, Appliances \cite{ni2019justifying}, Gift Cards \cite{ni2019justifying} & Accuracy, Precision, Recall, F1-score, MSE, RMSE, and MAE & Sentiment Classification on Streaming E-commerce Reviews \\
    & & \textbf{TAGrader}~\cite{pan2024distilling} & Cora \cite{mccallum2000automating}, PubMed \cite{sen2008collective}, ogbn-products \cite{hu2020open}, Arxiv-2023 \cite{he2024harnessing} & Accuracy & Node Classification on Text-attributed Graphs \\
    & & \textbf{SEGA}~\cite{chen2024depression} & DAIC-WOZ \cite{gratch2014distress}, EATD \cite{shen2022automatic} & Macro-F1 & Depression Detection \\
    & & \textbf{SocioHyperNet}~\cite{shu2024llm} & MBTI \cite{shu2024llm} & Accuracy, AUC, Macro-F1, Micro-F1, IMP & Examining Personality Traits \\
    & & \textbf{Cella}~\cite{zhang2024cost} & Cora \cite{mccallum2000automating}, Citeseer \cite{giles1998citeseer}, PubMed \cite{sen2008collective}, Wiki-cs \cite{mernyei2020wiki} & Accuracy, NMI, ARI, F1-score & Label‐free Node Classification \\
    & & \textbf{LLM-TIKG}~\cite{hu2024llm} & threat-dataset \cite{hu2024llm} & Precision, Recall, F1-Score & Threat Intelligence Knowledge Graph Construction \\
    & & \textbf{ANLM-assInNNER}~\cite{liao2025large} & NE dataset \cite{liao2025large} & Precision, Recall, F1-Score & Robotic Fault Diagnosis Knowledge Graph Construction \\
    & & \textbf{LLM-HetGDT}~\cite{ma2025llm} & Twitter-HetDrug \cite{ma2025llm} & Macro-F1, GMean & Online Drug Trafficking Detection \\
    \cmidrule(lr){2-6}
    & \multirow{4}{*}{Prediction}
     & \textbf{LLM-SBCL}~\cite{ni2024enhancing} & biology \cite{ni2024enhancing}, law \cite{ni2024enhancing}, cardiff20102 \cite{ni2024enhancing}, sydney19351 \cite{ni2024enhancing}, and sydney23146 \cite{ni2024enhancing} & Binary-F1, Micro-F1, Macro-F1, Accuracy & Student Performance Prediction \\
    &  & \textbf{LKPNR}~\cite{runfeng2023lkpnr} & MIND \cite{wu2020mind} & AUC, MRR, nDCG & Personalized News Recommendation \\
    & & \textbf{LLM-DDA}~\cite{gu2024empowering} & BCFR-dataset \cite{gu2024empowering} & AUC, AUPR, F1-score, Precision & Computational Drug Repositioning \\
    \cmidrule(lr){2-6}
    & \multirow{2}{*}{\makecell[c]{Graph \\Completion}}
      & \textbf{KICGPT}~\cite{wei2023kicgpt} &FB15k-237 \cite{toutanova2015observed}, WN18RR \cite{dettmers2018convolutional} &MRR, Hits@N & Link Completion \\
    & & \textbf{KGCD}~\cite{hou2025low} & WN18RR \cite{dettmers2018convolutional}, YAGO3-10 \cite{toutanova2015observed}, WN18 \cite{10.1145/3459637.3482263}& MRR, Hits@N & Low-resource Knowledge Graph Completion \\
    \cmidrule(lr){2-6}
    & {\makecell[c]{Foundation \\Model}}
      & \textbf{GraphCLIP}~\cite{zhu2024graphclip} & ogbn-arXiv \cite{hu2020open}, Arxiv-2023 \cite{he2024harnessing}, PubMed \cite{sen2008collective}, ogbn-products \cite{hu2020open}, Reddit \cite{huang2024can}, Cora \cite{mccallum2000automating}, CiteSeer \cite{giles1998citeseer}, Ele-Photo \cite{yan2023comprehensive}, Ele-Computers \cite{yan2023comprehensive}, Books-History \cite{yan2023comprehensive}, Wikics \cite{mernyei2020wiki}, Instagram \cite{huang2024can} & Accuracy & Transfer Learning on Text-attributed Graphs \\

    \midrule

    \multirow{6}{*}{\makecell[c]{Structure\\ Imbalance \\ Graph \\Learning}}
    & Node Classification
      & \textbf{GraphEdit}~\cite{guo2024graphedit} & Cora \cite{mccallum2000automating}, Citeseer \cite{giles1998citeseer}, PubMed \cite{sen2008collective} & Accuracy & Refining Graph Topologies \\
    \cmidrule(lr){2-6}
    & \multirow{3}{*}{\makecell[c]{Graph\\ Completion}}
      & \textbf{SATKGC}~\cite{ko2024subgraph} & WN18RR \cite{dettmers2018convolutional}, FB15k-237 \cite{toutanova2015observed}, Wikidata5M \cite{wang2021kepler} & MRR, Hits@N & Knowledge Graph Completion \\
    & & \textbf{MPIKGC}~\cite{xu2024multi} & FB15k-237 \cite{toutanova2015observed}, WN18RR \cite{dettmers2018convolutional}, FB13 \cite{socher2013reasoning}, WN11 \cite{socher2013reasoning} & MR, MRR, Hits@N, Accuracy & Knowledge Graph Completion \\
    & & \textbf{LLM4RGNN}~\cite{zhang2024can} & Cora \cite{mccallum2000automating}, Citeseer \cite{giles1998citeseer}, PubMed \cite{sen2008collective}, ogbn-arxiv \cite{hu2020open}, ogbn-products \cite{hu2020open} & Accuracy & Improving the Adversarial Robustness \\

    \bottomrule
  \end{tabular}
  \end{adjustbox}
  \label{tab:imbalance_methods}
\end{table*}

\subsubsection{Class-Imbalanced Graph Learning}
Class imbalance refers to scenarios where the sample size of certain categories is significantly smaller than others. For instance, in social networks, ordinary users vastly outnumber influential ``key opinion leaders'', while in fraud detection systems, legitimate transactions overwhelmingly dominate over fraudulent ones. Traditional classifiers, including GNNs, often exhibit a bias toward predicting majority-class samples, resulting in suboptimal recognition accuracy for minority classes \cite{ma2023class, liu2023survey}. Furthermore, GNNs update node representations by aggregating neighborhood information. Minority-class nodes risk being structurally homogenized or misrepresented when their local neighborhoods are dominated by majority-class nodes, leading to a propagation of representation bias across the graph.

\textbf{(1) Traditional Methods}
Traditional methods for addressing class imbalance in graph data primarily include clustering, contrastive learning, data augmentation, and innovative loss function. Most articles combine one or more of the above techniques to solve the class imbalance problem.

ECGN \cite{thapaliya2024ecgn} proposes a cluster-aware graph neural network framework that explicitly models the graph cluster structure to balance class representations, utilizing intra-cluster contrastive loss to enhance the distinguishability of minority class nodes. Similarly, C$^3$GNN \cite{ju2024cluster} clusters majority class nodes into subclasses and contrasts them with the minority class, using Mixup techniques to enhance the semantic substructure representation and alleviate the class imbalance problem. These methods share a key insight: clustering helps balance the class distribution by dividing majority class samples into multiple subclasses, and contrastive learning can improve the separability of minority nodes by constructing positive and negative sample pairs. Building on this, CCGC \cite{yang2023cluster} designs a clustering-based positive sample selection strategy and optimizes graph representations through hierarchical contrastive learning, reducing overfitting to the minority class. ImGCL \cite{zeng2023imgcl} introduces a progressive balanced sampling strategy, dynamically adjusting contrastive weights based on node centrality measures, and utilizes pseudo-labeling to address topological imbalance. These works highlight how clustering and contrastive learning can be synergistically combined to tackle both structural and label imbalance.

Another line of work focuses on data augmentation, which involves generating diverse topologies or attribute changes for minority classes to balance the data distribution. GraphSMOTE \cite{zhao2021graphsmote}, a pioneering method, interpolates minority nodes in the embedding space and predicts edges to preserve topological consistency. However, such interpolation-based methods may struggle with semantic coherence. To address this, GraphSHA \cite{li2023graphsha} generates hard samples via feature mixing and employs a semi-mixed connection strategy to prevent neighbor-class encroachment, enhancing boundary discrimination. Augmentation is often paired with adaptive loss functions. For example, TAM \cite{song2022tam} introduces topology-aware margin loss, dynamically adjusting classification boundaries based on local neighbor distributions of nodes, addressing decision bias caused by label node location shifts. 

Beyond data generation, another research direction emphasizes directly strengthening tail-class representations and rebalancing classifiers, often through retrieval or expert-based designs. For instance, RAHNet \cite{mao2023rahnet} combines a retrieval module with a second-stage classifier that applies max-norm and weight decay regularization. This joint design improves representation quality while reducing head-class dominance. Building on the idea of specialized modules, CoMe \cite{yi2023towards} introduces a collaborative-expert framework, where a general expert captures global shared patterns and a specific expert focuses on tail-class features; a collaboration regularizer further enforces complementarity, leading to more balanced performance. Extending this paradigm, KDEX \cite{mao2025learning} trains knowledge-diverse experts, each specializing in different structural or semantic patterns, and integrates their outputs via an expert allocation and fusion mechanism. Collectively, these approaches highlight how retrieval and expert-based architectures can complement generation strategies by directly enhancing the representation and robustness of tail classes.

In addition, HGIF \cite{ren2024heterophilic} addresses the heterogeneous graph scenario by using invariance learning to separate semantically invariant features, enhancing the robustness of fraud detection in out-of-distribution settings. GraphSANN \cite{liu2023imbalanced} breaks the homophily assumption and designs a topological reconstruction framework to tackle the class imbalance problem in heterogeneous graphs.

\begin{figure}[htbp]
\centerline{\includegraphics[width=0.95\linewidth]{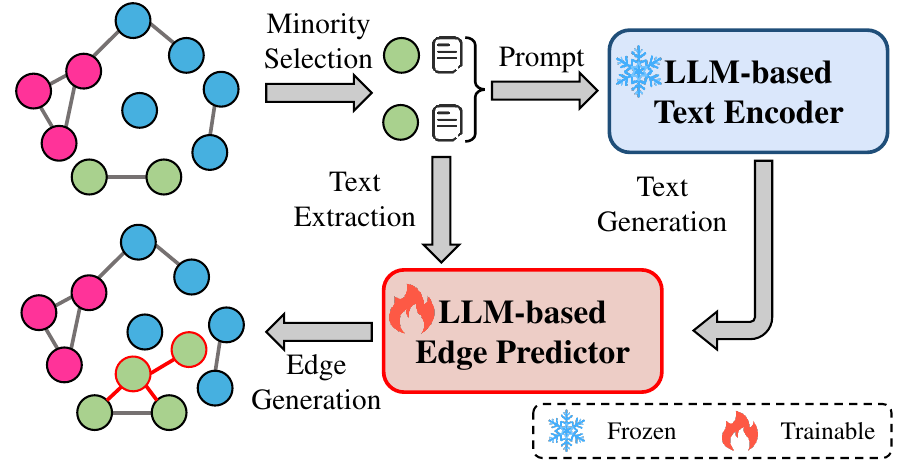}}
\caption{Illustration of an LLM-based pipeline for addressing class imbalance in graphs. Minority class information is first extracted and linearized into textual prompts, which are processed by the LLM to generate semantically enriched nodes and edges. The newly synthesized elements (highlighted in red) are integrated into the original graph to enhance representation and balance class distributions.}
\label{fig2-2}
\end{figure}

\textbf{(2) Methods Based on LLMs}
LLM-based techniques offer a novel approach to address class imbalance. Researchers can overcome the limitations of traditional methods by leveraging three aspects: semantic-driven data augmentation, context-aware text understanding, and external knowledge injection.  Figure \ref{fig2-2} illustrates how LLMs can address class imbalance in graph learning by synthesizing semantically rich representations for minority class nodes. In this framework, graph neighborhoods, node attributes, and type constraints are linearized into natural language prompts, serving as the sole input modality for the LLM. This allows the model to generate new nodes or edge candidates with both semantic coherence and structural compatibility.

First, some methods utilize the text generation capabilities of LLMs for node attribute descriptions (such as entity descriptions and label semantic expansion) to synthesize more semantically coherent features for minority class nodes. LA-TAG \cite{wang2024large} proposes a framework for text attribute rewriting and synthesis based on LLMs. It generates minority class node descriptions compatible with graph topology through instruction tuning, solving the semantic disconnection problem in text-graph alignment of traditional methods. Similarly, LLM4NG \cite{yu2025leveraging} designs a node generation approach for few-shot scenarios. It utilizes the LLM decoder to generate virtual nodes with both semantic coherence and structural consistency, while applying contrastive regularization to constrain the embedding space distribution. The method in \cite{zhang2024fine} also integrates graph structure understanding modules with LLM generators to achieve fine-grained synthesis of sparse data streams, dynamically balancing class distributions in real-time graph data.

Another approach focuses on the contextual understanding and reasoning capabilities of LLMs, mining implicit relationships of minority classes through structure-aware reasoning. Both LLM-GNN \cite{chen2023label} and Cella \cite{zhang2024cost} proposed a zero-shot classification framework based on LLM semantic reasoning. By parsing node attributes and neighbor context, they directly generate category prediction confidence and get rid of the dependence on labeled data. G2P2 \cite{wen2023augmenting} introduces a graph-guided pretraining prompt framework, using LLMs to generate augmented text related to graph structure, enhancing the discriminability of low-resource classes in heterogeneous spaces. Meanwhile, LLM-TIKG \cite{hu2024llm} builds a threat intelligence knowledge graph, using LLMs to extract attack pattern association rules from unstructured text, and enhancing the representation of minority class threat entities through heterogeneous graph attention.

To improve generalization on sparse or long-tail categories in graphs, another line of work focuses on injecting external knowledge—either from open-domain corpora or domain-specific sources—into the learning process. KICGPT \cite{wei2023kicgpt} enriches long-tail relational reasoning by dynamically constructing structure-aware prompts with contextual knowledge, significantly improving completion performance for sparse relations. In the recommendation domain, LKPNR \cite{runfeng2023lkpnr} integrates user preference reasoning from LLMs and semantic path mining to address cold-start scenarios in long-tail news recommendation. KGCD \cite{hou2025low} augments knowledge graph completion with pseudo-triplets generated from logical rules, guiding the model to focus on underrepresented relations. Other methods, such as \cite{pan2024distilling}, distill open-domain knowledge from LLMs into graph encoders through hierarchical knowledge transfer, while GraphCLIP \cite{zhu2024graphclip} leverages cross-modal alignment to construct graph–text contrastive learning objectives for tail-category enhancement. In biomedical applications, LLM-DDA \cite{gu2024empowering} infuses mechanistic knowledge into graph reconstruction, improving drug repurposing predictions for rare diseases. These works demonstrate that LLMs serve not just as predictors, but as external knowledge carriers that can guide graph learning systems beyond their native data distributions.

In addition to the aforementioned new technologies, the introduction of LLMs has accelerated innovation in cross-domain applications. SEGA \cite{chen2024depression} uses LLMs to parse the structured elements of clinical conversation graphs, extracting depression semantic clues and social interaction patterns, enhancing the recognition of minority positive samples in mental health classification. The method in \cite{shu2024llm} proposes an LLM-enhanced sociological analysis framework based on hypergraphs, addressing behavior class imbalance in social media data through personality trait hyperedge modeling and LLM semantic deconstruction. ANLM-assInNNER \cite{liao2025large} develops an automatic construction system for robot fault diagnosis knowledge graphs, using LLMs to generate fine-grained fault entity descriptions and balancing the data distribution of equipment status categories. LLM-SBCL \cite{ni2024enhancing} combines graph neural networks with LLM cognitive state modeling, improving the reliability of predicting questions on less common knowledge points through semantic enhancement of learner-question interaction graphs.

In summary, LLM-based approaches for class imbalance aim to even out the information content per class by generating additional data or highlighting distinguishing features, rather than just mathematically re-weighting or duplicating existing data. 

\subsubsection{Structure-Imbalanced Graph Learning}
Structural imbalance refers to skewness in graph topology that can hinder learning. A typical example is a hub node vs. peripheral node issue: Hub nodes with very high degree can dominate aggregation and also often have many more training signals, whereas low-degree nodes might be ignored or get a very noisy aggregate from a single neighbor. Similarly, certain substructures might be over-represented. For example, in a molecule graph dataset for drug discovery, maybe most molecules contain a benzene ring but only a few contain a rarer motif; a GNN might mostly learn features relevant to benzene rings and be less sensitive to the rare motif, which could be critical for certain properties.

\textbf{(1) Traditional Methods}
The main techniques for addressing structural imbalance in graph data can be broadly divided into two lines: debiasing-based approaches and structural enhancement-based approaches.  From the debiasing perspective, several methods focus on dynamically adjusting neighbor aggregation weights to reduce the over-dominance of high-centrality nodes. For example, DegFairGNN \cite{liu2023degfairgnn} introduces a generalized degree fairness constraint that reallocates neighborhood attention weights between high- and low-degree nodes, thereby mitigating bias in GNN message passing. Extending this idea to knowledge graphs, KG-Mixup \cite{shomer2023toward} analyzes degree bias in entity embeddings and proposes a degree-aware contrastive loss to balance geometric constraints between high- and low-degree entities. Together, these approaches highlight the importance of reweighting meachanisms for correcting degree-induced biases.  On the other hand, structural enhancement approaches aim to enrich the representation capacity of underrepresented nodes or substructures. SOLT-GNN \cite{liu2022size} improves graph classification by using a size-aware hierarchical pooling strategy that balances representation distributions through subgraph cropping and feature decoupling. Building on the idea of subgraph augmentation, SAILOR \cite{liao2023sailor} strengthens the visibility of low-degree nodes in the global topology via $k$-hop subgraph expansion and adversarial edge generation. Both methods demonstrate that explicitly enhancing substructures can effectively alleviate long-tailed imbalance in graphs.  A complementary line of work tackles imbalance from the perspective of global structure reconstruction and regularization. HiRe \cite{wu2022hierarchical} develops a hierarchical relational meta-learning framework with a meta-path-guided negative sampling mechanism, balancing the structural coverage density between head and tail relations in knowledge graphs. Similarly, QTIAH-GNN \cite{liu2023qtiah} introduces a heterogeneous GNN that jointly considers quantity imbalance and topological imbalance, using meta-relation-specific neighbor sampling together with topological entropy regularization. These methods illustrate how global structural modeling can address both numerical and topological imbalance.  

Although existing methods have alleviated the problem of graph structural imbalance to a certain extent, they still have limitations. They often rely on manually designed constraint rules, which leads to limited generalization ability and difficulty in adapting to complex and changing graph structural deviations. At the same time, these methods tend to ignore the semantic associations of the global topology, and usually adopt static adjustment strategies, which cannot dynamically adapt to the structural evolution patterns implicit in the graph \cite{ma2023class, liu2023survey}.

\begin{figure}[htbp]
\centerline{\includegraphics[width=0.95\linewidth]{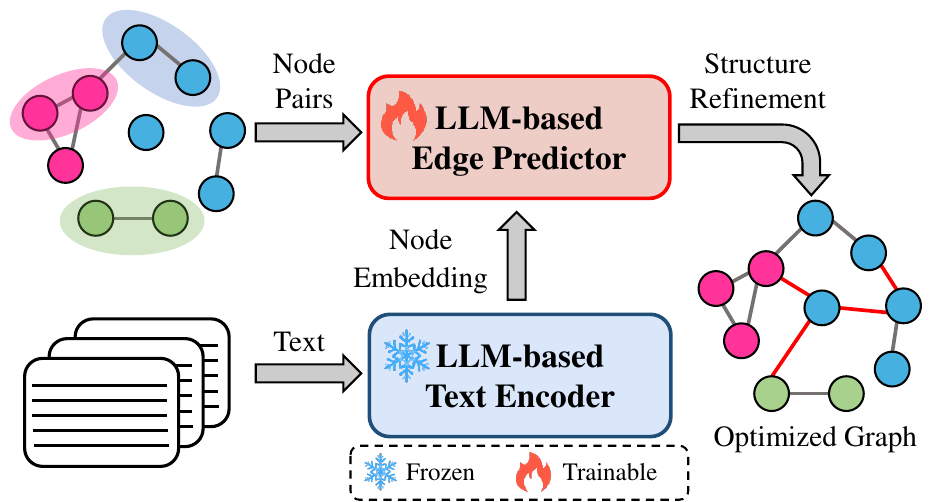}}
\caption{Structural Imbalance Remediation via LLM-Based Text-Topology Co-Optimization. The optimized graph (right) shows the rebalanced connectivity, with new connections indicated in red.}
\label{fig2-3}
\end{figure}

\textbf{(2) Methods Based on LLMs} LLM-based methods offer a new paradigm for addressing structural imbalance in graphs by generating semantically meaningful edits—such as context-aware virtual edges or subgraph structures—that enhance connectivity for long-tail nodes (Figure~\ref{fig2-3}). Unlike traditional GNN-based approaches, which learn structural patterns primarily from in-graph statistical distributions, LLM-based frameworks integrate rich prior semantic knowledge obtained from large-scale pretraining with graph-specific structures.

For example, SATKGC \cite{ko2024subgraph} constructs subgraph-aware embeddings and uses contrastive losses to reinforce coherence among tail relations. MPIKGC \cite{xu2024multi} integrates LLMs for semantic enhancement, structural correction, and logic-driven edge generation, effectively mitigating relational imbalance. LLM4RGNN \cite{zhang2024can} transfers GPT-4’s reasoning capacity to detect and recover from malicious structural edits in a lightweight graph refinement framework. Complementing these, GraphEdit \cite{guo2024graphedit} proposes an iterative generate-and-validate loop, where LLMs suggest candidate edges which are then filtered by topological validators to ensure semantic and structural consistency. These approaches demonstrate that, when graphs are represented in text-compatible forms, LLMs can inject new connections that are not arbitrary, but semantically aligned—offering principled structure augmentation for underrepresented classes.

\subsubsection{Evaluation for Handling Imbalance in Graphs}

We summarize the evaluation protocols for LLM-based methods addressing graph imbalance, covering key datasets, metrics, and downstream tasks (see Table~\ref{tab:imbalance_methods}). Commonly used datasets include Cora \cite{mccallum2000automating}, Citeseer~\cite{giles1998citeseer}, PubMed \cite{sen2008collective}, and ogbn-arxiv \cite{hu2020open} for node classification under class imbalance, and FB15k-237, WN18RR \cite{toutanova2015observed, dettmers2018convolutional} for knowledge graph completion in low-resource settings. Domain-specific benchmarks such as threat-dataset \cite{hu2024llm} and Twitter-HetDrug \cite{ma2025llm} are also adopted for specialized applications like fraud detection and mental health analysis.

Standard evaluation metrics vary by task: Accuracy and Macro-F1 are widely used for classification, while MRR and Hits@N are reported for link prediction and knowledge graph completion. For recommendation and retrieval tasks, AUC, MRR, and nDCG are common. Downstream tasks include few-shot node classification, sentiment analysis, knowledge graph construction, and drug repositioning, demonstrating the adaptability of LLMs in balancing both label and structural distributions. These evaluations highlight the effectiveness of LLM-based methods in enhancing model performance on underrepresented classes and nodes.

\subsubsection{Summary of Imbalance} Traditional graph imbalance methods mitigate bias via re-sampling, cost adjustment, or graph augmentation, but they lack external semantic context. LLM-integrated methods aim to generate new graph content or features that specifically bolster the minority classes or structures. As reported in recent studies, using LLMs in this way can significantly improve classification performance on long-tail classes and ensure that even structurally unique nodes are recognized. Essentially, LLMs function as an intelligent oversampling mechanism: instead of naive duplication, they produce novel yet relevant samples in data space. The result is often more balanced training data for the graph model and better generalization to minority cases.

\subsection{Cross-Domain Heterogeneity in Graphs}

Real-world graph data is often collected from multiple source domains \cite{petermann2014graph, collarana2017semantic, li2023multi}, which can exhibit significant heterogeneity, referring to extreme disparities in both attributes and structural patterns. This heterogeneity typically arises when the graph data is collected or integrated from domains with inconsistent data modalities or distributions. Cross-domain heterogeneity introduces significant challenges for graph analysis and modeling, as the severe disparities make it difficult to unify these data into a common representation space \cite{zhang2019dane, tsai2016learning} and even hinder the detection of valuable and transferable features that can generalize well throughout the graph \cite{hassani2022cross, zhu2021shift}. Moreover, addressing cross-domain heterogeneity is a key prerequisite for building graph foundation models for more generalizable and scalable graph learning across diverse real-world applications \cite{liu2024one}. 

While traditional graph domain adaptation methods \cite{wu2024graph} and multimodal graph learning methods \cite{peng2024learning} can help bridge the gap between different distributions or modalities from different graph domains, they typically require large-scale training datasets from different domains and cannot generalize to some unseen domains during training. In contrast, LLMs, with their superior comprehension and generalization abilities, can extract valuable and transferable semantic and structural features within a unified representation space without training from scratch, effectively addressing the challenges posed by cross-domain heterogeneity in graph learning \cite{liu2024revisiting}. 

\begin{figure}[t]
\centerline{\includegraphics[width=0.9\linewidth]{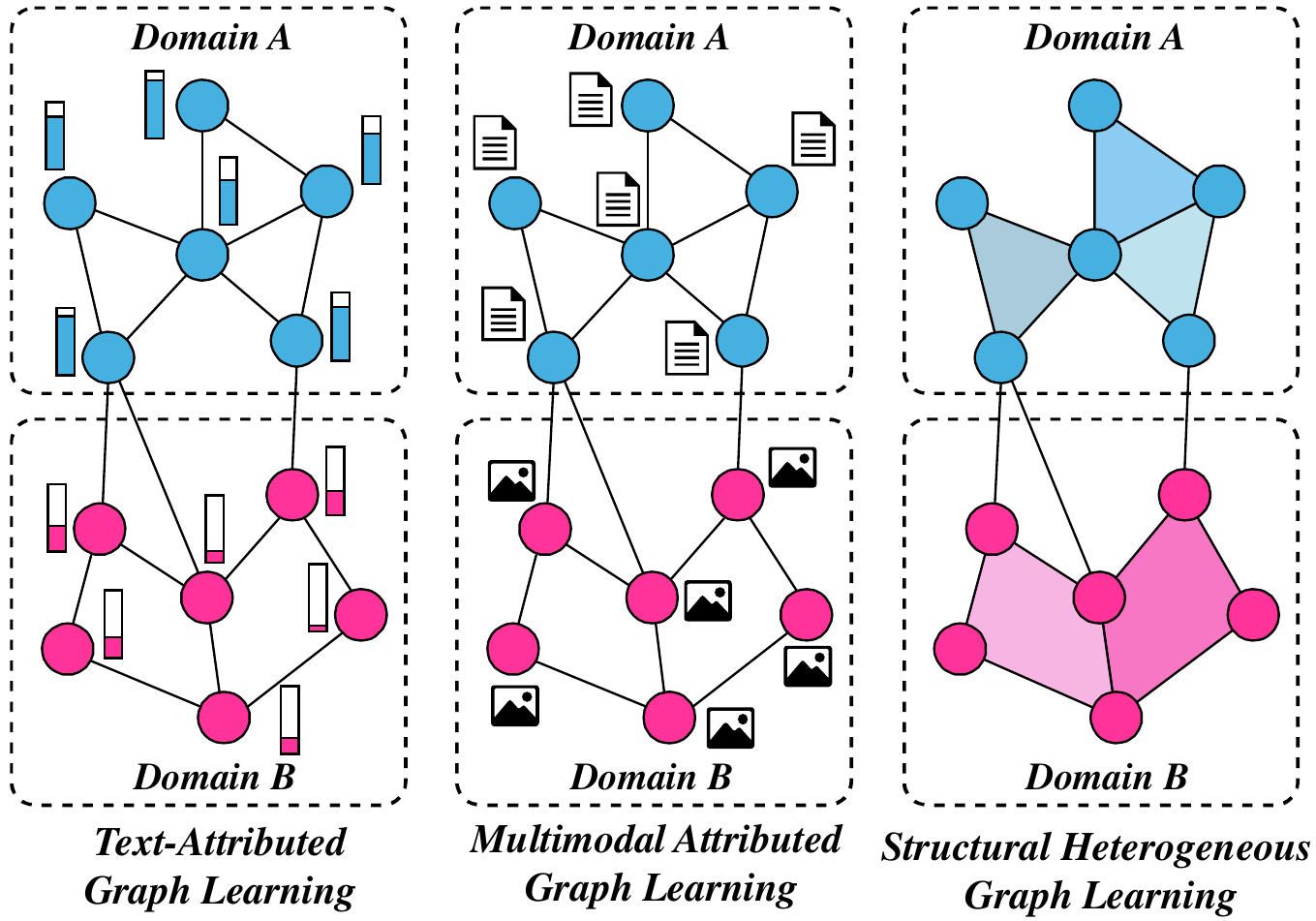}}
\caption{Illustration of three representative tasks addressing cross-domain heterogeneity in graphs. Text-attributed graph learning (left) involves graphs where nodes or edges are associated with textual content, requiring joint modeling of structure and language.
Multimodal attributed graph learning (middle) integrates diverse data types across nodes or relations.
Structural heterogeneous graph learning (right) with complex cross-modal dependencies and heterogeneous attribute spaces.}
\label{fig:cross_domain_heterogeneity}
\end{figure}

In this section, as illustrated in Figure~\ref{fig:cross_domain_heterogeneity}, we categorize cross-domain heterogeneity in graph data into three types: within-modality attribute heterogeneity, cross-modality attribute heterogeneity, and structural heterogeneity. Since LLMs are specifically designed for modeling textual modality, our focus on within-modality attribute heterogeneity primarily centers around textual attributes. For each type, we first review traditional graph learning methods, analyzing their strengths and limitations, and then discuss recent research leveraging LLMs to unify graph data and address the corresponding challenges. The relevant references and categorization are summarized in Table~\ref{tab:references_heterogeneity_grouped}.

\begin{table*}[htbp]\scriptsize
  \centering
  \caption{LLM-based methods for handling \textbf{cross-domain heterogeneity} in graphs, grouped by domains and tasks, with representative methods, datasets, metrics, and downstream tasks.}
  \renewcommand{\arraystretch}{2}
  \setlength{\tabcolsep}{3.5pt}
  \begin{adjustbox}{max width = 1.0\linewidth}
  \begin{tabular}{c P{0.1\linewidth} P{0.1\linewidth} P{0.3\linewidth} P{0.15\linewidth} P{0.15\linewidth}}
    \toprule
    \textbf{Domains} & \textbf{Tasks} & \textbf{Methods} & \textbf{Typical Datasets} & \textbf{Common Metrics} & \textbf{Downstream Tasks} \\
    \midrule

    \multirow{13}{*}{\makecell[c]{Text-\\Attributed\\Graph\\Learning}}
    & \multirow{4}{*}{\makecell[c]{Textual\\Attribute\\Alignment}}
      & \textbf{TAPE}~\cite{he2024harnessing} & Cora~\cite{mccallum2000automating}, PubMed~\cite{sen2008collective}, Arxiv-2023~\cite{he2024harnessing}, ogbn-arxiv~\cite{hu2020open}, ogbn-products~\cite{hu2020open} & Accuracy & Node Classification \\
    & & \textbf{LLMRec}~\cite{wei2024llmrec} & MovieLens~\cite{harper2015movielens}, Netflix~\cite{netflix_prize_data} & Recall, NDCG, Precision & Item Recommendation \\
    & & \textbf{MINGLE}~\cite{cui2024multimodal} & MIMIC-III~\cite{johnson2016mimic}, CRADLE~\cite{cui2024multimodal} & Accuracy, AUC, AUPR, F1 & Node Classification\\
    & & \textbf{GHGRL}~\cite{gao2024bootstrapping} & IMDB~\cite{zhang2019heterogeneous}, DBLP~\cite{zhang2019heterogeneous}, ACM~\cite{zhang2019heterogeneous}, Wiki-CS~\cite{mernyei2020wiki}, IMDB-RIR~\cite{gao2024bootstrapping}, DBLP-RID~\cite{gao2024bootstrapping} & Macro-F1, Micro-F1 & Node Classification \\
    \cmidrule(lr){2-6}
    & \multirow{7}{*}{\makecell[c]{Graph\\Foundation\\Model}}
      & \textbf{OFA}~\cite{liu2024one} & Cora~\cite{mccallum2000automating}, PubMed~\cite{sen2008collective}, ogbn-arxiv~\cite{hu2020open}, Wiki-CS~\cite{mernyei2020wiki}, MOLHIV~\cite{wu2018moleculenet}, MOLPCBA~\cite{wu2018moleculenet}, FB15K237~\cite{toutanova2015observed}, WN18RR~\cite{dettmers2018convolutional}, ChEMBL~\cite{gaulton2012chembl} & Accuracy, AUC, AUPR & Node Classification, Link Prediction, Graph Classification \\
    & & \textbf{UniGraph}~\cite{he2024unigraph} & Cora~\cite{mccallum2000automating}, PubMed~\cite{sen2008collective}, ogbn-arxiv~\cite{hu2020open}, ogbn-products~\cite{hu2020open}, Wiki-CS~\cite{mernyei2020wiki}, FB15K237~\cite{toutanova2015observed}, WN18RR~\cite{dettmers2018convolutional}, MOLHIV~\cite{wu2018moleculenet}, MOLPCBA~\cite{wu2018moleculenet} & AUC & Node Classification, Link Prediction, Graph Classification \\
    & & \textbf{BooG}~\cite{cheng2024boosting} & Cora~\cite{mccallum2000automating}, PubMed~\cite{sen2008collective}, ogbn-arxiv~\cite{hu2020open}, Wiki-CS~\cite{mernyei2020wiki}, MOLHIV~\cite{wu2018moleculenet}, MOLPCBA~\cite{wu2018moleculenet} & AUC & Node Classification, Graph Classification \\
    & & \textbf{Hyper-FM}~\cite{feng2025hypergraph} & Cora-CA-Text~\cite{feng2025hypergraph}, Cora-CC-Text~\cite{feng2025hypergraph}, Pubmed-CA-Text~\cite{feng2025hypergraph}, Pubmed-CC-Text~\cite{feng2025hypergraph}, AminerText~\cite{feng2025hypergraph}, Arxiv-Text~\cite{feng2025hypergraph}, Movielens-Text~\cite{feng2025hypergraph}, IMDB-Text~\cite{feng2025hypergraph}, GoodBook-Text~\cite{feng2025hypergraph}, PPI-Text~\cite{feng2025hypergraph} & Accuracy & Node Classification \\

    \midrule

    \multirow{12}{*}{\makecell[c]{Multimodal\\Attributed\\Graph\\Learning}}
    & \multirow{3}{*}{\makecell[c]{MLLM-based\\Multimodal\\Alignment}}
      & \textbf{LLMRec}~\cite{wei2024llmrec} & MovieLens~\cite{harper2015movielens}, Netflix~\cite{netflix_prize_data} & Recall, NDCG, Precision & Item Recommendation \\
    & & \textbf{MAGB}~\cite{yan2024when} & Cora~\cite{mccallum2000automating}, Wiki-CS~\cite{mernyei2020wiki}, Ele-Photo~\cite{yan2023comprehensive}, Flickr~\cite{zenggraphsaint}, Movies~\cite{yan2024when}, Toys~\cite{yan2024when}, Grocery~\cite{yan2024when}, Reddit-S~\cite{yan2024when}, Reddit-M~\cite{yan2024when} & Accuracy, F1 & Node Classification \\
    \cmidrule(lr){2-6}
    & \multirow{8}{*}{\makecell[c]{Graph-Enhanced\\Multimodal\\Alignment}}
      & \textbf{MMGL}~\cite{yoon2023multimodal} & WikiWeb2M~\cite{burns2023suite} & BLEU-4, ROUGE-L, CIDEr & Section Summarization \\
    & & \textbf{GraphAdapter}~\cite{li2024graphadapter} & ImageNet~\cite{deng2009imagenet}, StandfordCars~\cite{krause20133d}, UCF101~\cite{soomro2012ucf101}, Caltech101~\cite{fei2004learning}, Flowers102~\cite{nilsback2008automated}, SUN397~\cite{xiao2010sun}, DTD~\cite{cimpoi2014describing}, EuroSAT~\cite{helber2019eurosat}, FGVCAircraft~\cite{maji2013fine}, OxfordPets~\cite{parkhi2012cats}, Food101~\cite{bossard2014food} & Accuracy & Image Classification \\
    & & \textbf{TouchUp-G}~\cite{zhu2024touchup} & ogbn-arxiv~\cite{hu2020open}, ogbn-products~\cite{hu2020open}, Books~\cite{zhu2024touchup}, Amazon-CP~\cite{zhu2024touchup} & MRR, Hits@N, Accuracy & link prediction, node classification \\
    & & \textbf{UniGraph2}~\cite{he2025unigraph2} & Cora~\cite{mccallum2000automating}, PubMed~\cite{sen2008collective}, ogbn-arxiv~\cite{hu2020open}, ogbn-papers100M~\cite{hu2020open}, ogbn-products~\cite{hu2020open}, Wiki-CS~\cite{mernyei2020wiki}, FB15K237~\cite{toutanova2015observed}, WN18RR~\cite{dettmers2018convolutional}, Amazon-Sports~\cite{zhu2024multimodal}, Amazon-Cloth~\cite{zhu2024multimodal}, Goodreads-LP~\cite{zhu2024multimodal}, Goodreads-NC~\cite{zhu2024multimodal}, Ele-Fashion~\cite{zhu2024multimodal}, WikiWeb2M~\cite{burns2023suite} & Accuracy, BLEU-4, ROUGE-L, CIDEr & Node Classification, Edge Classification, Section Summarization \\

    \midrule

    \multirow{10}{*}{\makecell[c]{Structural\\Heterogeneous\\Graph\\Learning}}
    & \multirow{10}{*}{\makecell[c]{Topological\\Graph\\Textualization}}
      & \textbf{LLMtoGraph}~\cite{liu2023evaluating} & synthetic graph data~\cite{liu2023evaluating} & Accuracy, Positive Response Ratio & Node Classification, Path Finding, Pattern Matching \\
    & & \textbf{NLGraph}~\cite{wang2023can} & NLGraph~\cite{wang2023can} & Accuracy, Partial Credit Score, Relative Error & Path Finding, Pattern Matching, Topological Sort \\
    & & \textbf{Talk-like-a-Graph}~\cite{fatemi2024talk} & GraphQA~\cite{fatemi2024talk} & Accuracy & Link Prediction, Pattern Matching \\
    & & \textbf{GPT4Graph}~\cite{guo2023gpt4graph} & ogbn-arxiv~\cite{hu2020open}, MOLHIV~\cite{wu2018moleculenet}, MOLPCBA~\cite{wu2018moleculenet}, MetaQA~\cite{zhang2018variational} & Accuracy & Node Classification, Graph Classification, Graph Query Language Generation \\
    & & \textbf{GITA}~\cite{wei2024gita} & GVLQA~\cite{wei2024gita} & Accuracy & Link Prediction, Pattern Matching, Path Finding, Topological Sort \\
    & & \textbf{LLM4-Hypergraph}~\cite{feng2025beyond} & LLM4Hypergraph~\cite{feng2025beyond} & Accuracy & Isomorphism Recognition, Structure Classification, Link Prediction, Path Finding \\

    \bottomrule
  \end{tabular}
  \end{adjustbox}
  \label{tab:references_heterogeneity_grouped}
\end{table*}

\begin{table*}[htbp]\scriptsize
  \centering
  \addtocounter{table}{-1}
  \caption{LLM-based methods for handling \textbf{cross-domain heterogeneity} in graphs, grouped by domains and tasks, with representative methods, datasets, metrics, and downstream tasks. (continued from previous page)}
  \renewcommand{\arraystretch}{1.8}
  \setlength{\tabcolsep}{3.5pt}
  \begin{adjustbox}{max width = 1.0\linewidth}
  \begin{tabular}{c P{0.1\linewidth} P{0.1\linewidth} P{0.3\linewidth} P{0.15\linewidth} P{0.15\linewidth}}
    \toprule
    \textbf{Domains} & \textbf{Tasks} & \textbf{Methods} & \textbf{Typical Datasets} & \textbf{Common Metrics} & \textbf{Downstream Tasks} \\
    \midrule

    \multirow{15}{*}{\makecell[c]{Structural\\Heterogeneous\\Graph\\Learning}}
    & \multirow{9}{*}{\makecell[c]{Attributed\\Graph\\Textualization}} & \textbf{GraphText}~\cite{zhao2023graphtext} & Cora~\cite{mccallum2000automating}, Citeseer~\cite{giles1998citeseer}, Texas~\cite{pei2020geom}, Wisconsin~\cite{pei2020geom}, Cornell~\cite{pei2020geom} & Accuracy & Node Classification \\
    & & \textbf{WalkLM}~\cite{tan2024walklm} & PubMed~\cite{sen2008collective}, MIMIC-III~\cite{johnson2016mimic} & Macro-F1, Micro-F1, AUC, MRR & Node Classification, Link Prediction \\
    & & \textbf{Path-LLM}~\cite{shang2024path} & Cora~\cite{mccallum2000automating}, Citeseer~\cite{giles1998citeseer}, PubMed~\cite{sen2008collective}, ogbn-arxiv~\cite{hu2020open} & Macro-F1, Micro-F1, AUC, Accuracy & Node Classification, Link Prediction \\
    & & \textbf{InstructGLM}~\cite{ye2024language} & Cora~\cite{mccallum2000automating}, PubMed~\cite{sen2008collective}, ogbn-arxiv~\cite{hu2020open} & Accuracy & Node Classification, Link Prediction \\
    & & \textbf{MuseGraph}~\cite{tan2024musegraph} & Cora~\cite{mccallum2000automating}, ogbn-arxiv~\cite{hu2020open}, MIMIC-III~\cite{johnson2016mimic}, AGENDA~\cite{koncel2019text}, WebNLG~\cite{gardent2017creating} & Macro-F1, Micro-F1, Weighted-F1, BLEU-4, METEOR, ROUGE-L, CHRF++ & Node Classification, Graph-to-Text Generation \\
    & & \textbf{Graph-LLM}~\cite{chen2024exploring} & Cora~\cite{mccallum2000automating}, Citeseer~\cite{giles1998citeseer}, PubMed~\cite{sen2008collective}, ogbn-arxiv~\cite{hu2020open}, ogbn-products~\cite{hu2020open} & Accuracy & Node Classification \\
    \cmidrule(lr){2-6}
    & \multirow{7}{*}{\makecell[c]{Graph\\Token\\Learning}}
      & \textbf{GNP}~\cite{tian2024graph} & OBQA~\cite{mihaylov2018can}, ARC~\cite{clark2018think}, PIQA~\cite{bisk2020piqa}, Riddle~\cite{lin2021riddlesense}, PQA~\cite{jin2019pubmedqa}, BioASQ~\cite{tsatsaronis2015overview} & Accuracy & Question Answering \\
    & & \textbf{GraphToken}~\cite{perozzi2024let} & GraphQA~\cite{fatemi2024talk} & Accuracy & Link Prediction, Pattern Matching \\
    & & \textbf{GraphGPT}~\cite{tang2024graphgpt} & Cora~\cite{mccallum2000automating}, PubMed~\cite{sen2008collective}, ogbn-arxiv~\cite{hu2020open} & Accuracy, Macro-F1, AUC & Node Classification, Link Prediction \\
    & & \textbf{LLaGA}~\cite{chen2024llaga} & Cora~\cite{mccallum2000automating}, PubMed~\cite{sen2008collective}, ogbn-arxiv~\cite{hu2020open}, ogbn-products~\cite{hu2020open} & Accuracy & Node Classification, Link Prediction \\
    & & \textbf{TEA-GLM}~\cite{wang2024llms} & Cora~\cite{mccallum2000automating}, PubMed~\cite{sen2008collective}, ogbn-arxiv~\cite{hu2020open}, TAG benchmark~\cite{yan2023comprehensive} & Accuracy, AUC & Node Classification, Link Prediction \\
    & & \textbf{HiGPT}~\cite{tang2024higpt} & IMDB~\cite{zhang2019heterogeneous}, DBLP~\cite{zhang2019heterogeneous}, ACM~\cite{zhang2019heterogeneous} & Macro-F1, Micro-F1, AUC & Node Classification \\

    \bottomrule
  \end{tabular}
  \end{adjustbox}
\end{table*}

\subsubsection{Text-Attributed Graph Learning}
Textual attributes are common in real-world graph data, such as the abstract of each paper in a co-citation network \cite{yang2021graphformers} or item descriptions in a recommendation network \cite{wei2024llmrec}. Although these textual attributes appear in the same modality, they can demonstrate significant distribution heterogeneity due to different sources. For example, textual attributes such as clinical notes could come from different healthcare providers, each with their own writing style, vocabulary, and context \cite{mashima2024information}. Similarly, clinical notes might be informal and easy to understand for patients, while medical codes should be more formal and standardized. Moreover, the textual attributes can appear in different languages \cite{moreo2022generalized}, further increasing the heterogeneity. The distribution heterogeneity in textual attributes can lead to inconsistent semantic representations, making it difficult for graph learning methods to effectively capture transferable features.

\textbf{(1) Traditional Methods for Textual Attribute Modeling} 
To handle the heterogeneity in textual data, early data integration methods extract structured information from unstructured textual inputs, thereby reducing variability across sources and producing unified features for downstream tasks \cite{ford2016extracting, dong2018data, sheikhalishahi2019natural}. However, these approaches require manual design of data schemes and training extraction models tailored to specific applications. In contrast, statistical methods, such as Bag of Words and TF-IDF \cite{sparck1972statistical}, have been introduced to automatically generate unified feature vectors from text without relying on domain-specific design. These feature vectors are often used as inputs for graph learning methods, such as GNNs \cite{kipf2017semisupervised, hamilton2017inductive, velivckovic2018graph}, which further incorporate structural information from the graph to obtain more effective node representations for addressing downstream tasks. For instance, TADW \cite{yang2015network} approximates DeepWalk \cite{perozzi2014deepwalk} using matrix factorization, where the TF-IDF feature vectors of node textual attributes serve as the initial feature matrix. Paper2vec \cite{ganguly2017paper2vec} utilizes learnable Word2vec \cite{mikolov2013distributed} text embeddings as initial node features, which are then trained by predicting whether two nodes belong to the same neighborhood in the graph. While these early statistical methods or shallow models offer solutions for mapping heterogeneous textual attributes into a unified space, their limited expressiveness cannot effectively capture complex features in these textual attributes. Moreover, these methods adopt text embeddings as fixed initial node features rather than integrating them with structural learning, limiting the potential of leveraging graph topology to enhance the textual attributes representation learning \cite{yang2021graphformers}.

Several advanced methods explore approaches for modeling textual attributes that can better integrate with graph structures, aiming to improve the representation learning of both textual attributes and graph structure data simultaneously. Some approaches, like GIANT~\cite{chien2022node} and its extension E2EG~\cite{dinh2023e2eg}, leverage self-supervised learning to force text embeddings to encode graph-dependent information, which is trained by predicting a node's neighbors from its text. Other methods focus on creating joint architectures. For example, GraphFormers \cite{yang2021graphformers} proposes a GNN-nested architecture where the transformer-based textual embedding modules and the GNN modules are nested and trained together. Heterformer \cite{jin2023heterformer} introduces virtual neighbor tokens that capture information from both text-rich and textless neighbors, allowing the model to simultaneously consider textual semantic information and structural information to embed the textual node attributes. To capture relation-specific signals in heterogeneous graphs, METAG \cite{jin2023learning} employs a single pretrained language model to learn multiplex text representations, which can effectively handle the diverse semantic relations while maintaining high parameter efficiency.

Although these advanced methods can embed heterogeneous textual node attributes into a unified representation space, their dependence on complex deep architectures requires a sufficient amount of training data. In applications with limited data availability, the model may fail to capture the valuable features from excessive heterogeneous textual attributes, and the learned features might fail to generalize effectively to textual attributes from other sources. 

\textbf{(2) LLM-based Methods for Textual Attribute Modeling} LLMs, as language models, are naturally suited for modeling attributes represented in textual modality. With advanced language comprehension and generalization capabilities, LLMs can effectively capture the semantic meanings from heterogeneous textual attributes across diverse source domains and project them into a unified representation space that preserves the semantic information. 

\begin{figure}[t]
\centerline{\includegraphics[width=0.95\linewidth]{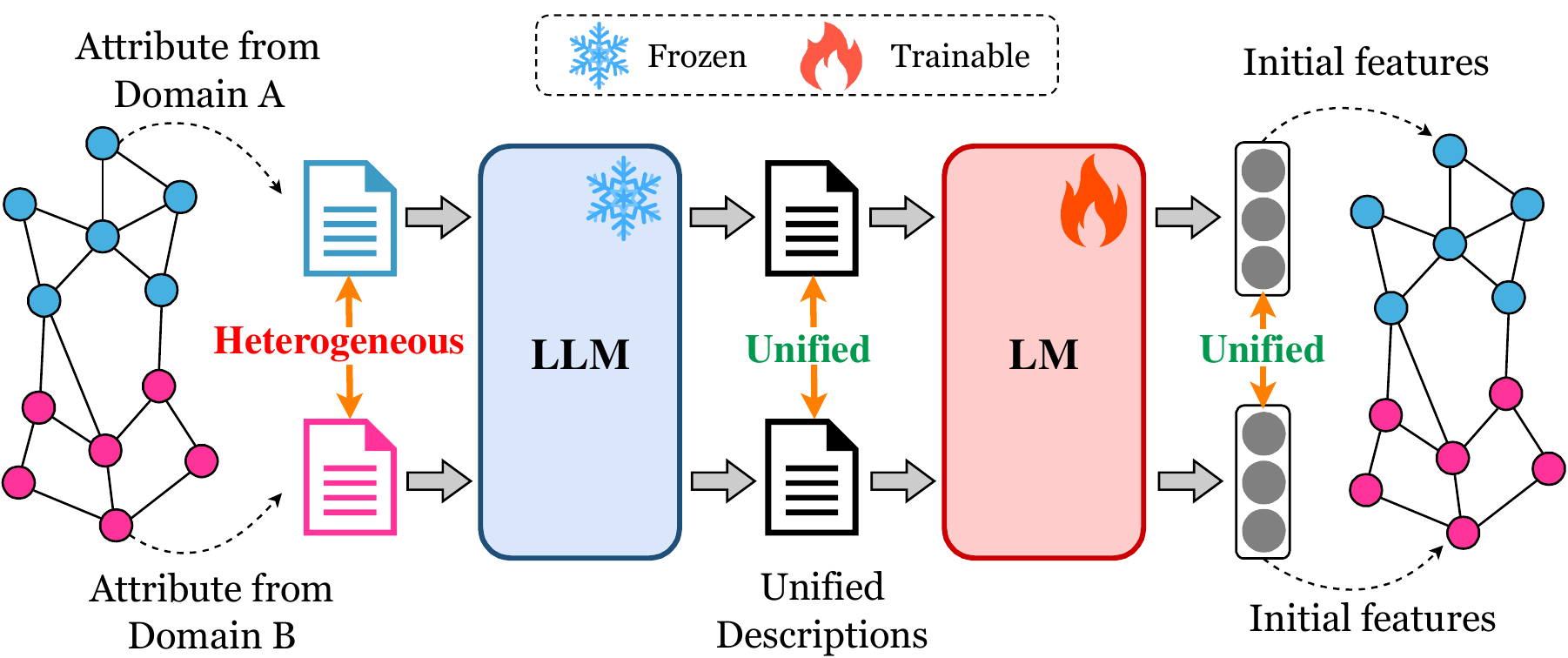}}
\caption{Handling cross-domain textual attribute heterogeneity with LLMs. Attributes from different domains are first transformed into unified textual descriptions by an LLM, then converted into unified embeddings using a trainable small LM to generate initial features for graph learning methods.}
\label{fig:llm_for_textual_attribute_heterogeneity}
\end{figure}

A number of methods leverage the use of LLMs to generate unified textual descriptions, which can be fed into a trainable smaller language model to extract task-specific embeddings, as depicted in Figure~\ref{fig:llm_for_textual_attribute_heterogeneity}. TAPE \cite{he2024harnessing} leverages the powerful language comprehension capabilities of LLMs to predict the category of a node based on its textual attributes and generate explanations for the prediction, which can be regarded as enhanced and aligned textual outputs. These aligned textual outputs are fed into a smaller language model to generate node feature vectors. As shown in \cite{liu2024revisiting}, the domain shift is reduced in the LLM-enhanced text compared to the original textual attributes. Similarly, LLMRec \cite{wei2024llmrec} leverages LLMs to summarize user profiles and item attributes within a recommendation network, reducing the heterogeneity of the original attributes and producing aligned textual representations. These representations are then reprocessed by LLMs to generate unified embedding vectors for both users and items. Instead of generating unified textual descriptions, MINGLE \cite{cui2024multimodal} utilizes LLMs to map clinical notes and medical codes into a unified embedding space and uses these unified embedding vectors as initial node features for training a hypergraph neural network. GHGRL \cite{gao2024bootstrapping} utilizes LLMs to unify textual attributes on heterogeneous graphs based on format types and content types. Specifically, it first summarizes types information of all nodes using LLMs, which are then used to generate attribute summaries for each node based on the predicted types. These generated attribute summaries are then fed into a language model to generate unified feature vectors for training a graph neural network. 

By transforming categorical and numerical features into textual attributes with language-based descriptions, LLMs can extend their ability to handle the heterogeneity in these structured attributes, which do not naturally belong to the textual modality. For example, in molecular graph modeling, node attributes such as atom types or properties are typically represented as categorical or numerical values. By converting these attributes into descriptive text and enriching with domain-specific explanations, LLMs can effectively understand their semantic meaning and unify them within a common representation space. Both OFA \cite{liu2024one} and UniGraph \cite{he2024unigraph} address attribute heterogeneity by constructing text-attributed graphs. These models enhance textual attributes by incorporating additional semantically rich contextual descriptions. Furthermore, they leverage domain knowledge to enrich non-textual attributes with textual representations, enabling LLMs to process diverse attributes in a unified manner and generate consistent embedding vectors. BooG \cite{cheng2024boosting} follows a similar approach but further employs a contrastive learning-based pretraining objective, which enhances the ability to learn expressive representations and generalize across different domains and downstream tasks. Similarly, Hyper-FM \cite{feng2025hypergraph} leverages a language model to extract semantic features from cross-domain textual attributes on hypergraphs and integrates structural information through hierarchical high-order neighbor prediction. 

These LLM-based methods utilize LLMs to comprehend semantic information from heterogeneous textual attributes and generate unified textual descriptions or vector embeddings for downstream graph learning methods. The unified and high-quality outputs from LLMs greatly simplify the downstream learning process and enhance performance by addressing the challenges posed by cross-domain heterogeneity in textual attributes.

\subsubsection{Multimodal Attributed Graph Learning}
Compared to attributes within the same modality, multimodal attributes encompass information from various modalities, such as text, images, audio, and videos. These multimodal attributes offer a more comprehensive and enriched representation of the underlying data. By capturing different aspects of a node through diverse inputs from different sources, multimodal attributes enhance the contextual understanding and provide complementary information that single-modal data may lack \cite{hu2021graph, wang2020multimodal, wei2019mmgcn}. However, this diversity also introduces additional challenges, as attribute heterogeneity across different domains manifests not only in data distribution but also in data formats. This variability complicates the unified processing of these attributes, requiring effective fusion methods to integrate information across different modalities \cite{yan2024when, wilcke2020end}.

\textbf{(1) Traditional Methods for Multimodal Attribute Modeling} 
Traditional methods aim to learn alignment or fusion patterns across different modalities by training models to capture the underlying relationships between heterogeneous multimodal attributes. MMGCN \cite{wei2019mmgcn} learns a multimodal graph convolution network on a user-item bipartite graph to learn modal-specific representations of users and micro-videos to better capture user preferences in different modalities. scMoGNN \cite{wen2022graph} utilizes different GNNs to learn representations of each cell-feature bipartite graph in different modalities, and finally concatenates these representations to fuse information from different modalities. \cite{cai2022graph} and \cite{he2023multimodal} introduce an attention mechanism in graph neural networks to dynamically capture the importance of information from different modalities. To reduce the need for large amounts of labeled data and improve robustness, contrastive learning methods are proposed for modeling multimodal graph data. Joyful \cite{li2023joyful} designs a global contextual fusion module and a specific modalities fusion module to capture information at different scales, and then concatenates the representations from both modules to generate a unified representation vector for each node. These modules are trained by comparing positive and negative pairs in the corrupted graphs through edge perturbation and random masking. FormNetV2 \cite{lee2023formnetv2} proposes a centralized multimodal graph contrastive learning strategy to learn fused representations from different modalities in one loss. HGraph-CL \cite{lin2022modeling} introduces a hierarchical graph contrastive learning framework that builds intra-modal and inter-modal graphs, leveraging graph augmentations and contrastive learning to capture complex sentiment relations within and across different modalities. 

While these traditional methods can align or fuse multimodal attributes in graph data, they still rely on abundant data to train the model and cannot easily generalize to an unseen domain. Compared to LLMs, these traditional methods are less adaptable to diverse data distributions and struggle to leverage pretrained knowledge for more efficient generalization.

\textbf{(2) LLM-based Methods for Multimodal Attribute Modeling} Although LLMs are specifically designed for understanding natural languages and not naturally suited for handling multimodal attributes, they can be combined or aligned with models for other modalities. Leveraging their superior generalization ability, LLM-based methods eliminate the need for large amounts of application-specific training data, opening up new possibilities for addressing attribute heterogeneity across multimodal source domains. 

Recent research on MLLMs focuses on developing advanced methods to align LLMs with models from other modalities \cite{lyu2024unibind, sun2024emu, li2023blip, liu2024visual, wang2024modaverse}.  Although these methods are not specifically designed for graph data, they can be used as powerful preprocessing tools for aligning multimodal attributes and generating unified representations. For example, by unifying textual and visual side information through a pretrained model Clip-ViT \cite{radford2021learning}, LLMRec \cite{wei2024llmrec} effectively mitigates the multimodal heterogeneity and enhances the node features in the recommendation network, leading to significant performance improvements. MAGB \cite{yan2024when} conducts experiments on a set of large multimodal attributed graph datasets, which demonstrate that MLLMs can effectively alleviate the biases from cross-domain multimodal heterogeneity. Recent approaches align encoders for different modalities by leveraging graph structures, making them more naturally suited for graph learning tasks. MMGL \cite{yoon2023multimodal} utilizes LLMs and image encoders with adapter layers to embed text and image attributes, respectively. These embeddings are then combined with graph positional encodings to capture graph structure information and finally fed into LLMs to generate the corresponding outputs. GraphAdapter \cite{li2024graphadapter} introduces GNN-based adaptors for encoders of different modalities, which can better align these encoders based on the graph structure information. TouchUp-G \cite{zhu2024touchup} improves node features obtained from pretrained models of different modalities by adapting them to the graph structure, using a new metric called feature homophily to quantify the correlation between the graph and node features, which enhances GNN performance across different tasks and data modalities. Following this direction, UniGraph2 \cite{he2025unigraph2} leverages modality-specific encoders alongside a GNN and an MoE module to effectively unify multimodal features while preserving the underlying graph structure.

By aligning LLMs with models for different modalities, these methods can simultaneously understand the heterogeneity across modalities and map the multimodal attributes into a unified embedding space. These unified attribute representations simplify the downstream graph learning process and improve the performance of multi-modal graph learning tasks. 

\subsubsection{Structural Heterogeneous Graph Learning}

Graph structures capture essential connectivity patterns that are crucial for real-world applications. However, graphs constructed from heterogeneous source domains can exhibit excessive heterogeneity in structural patterns. The structural heterogeneity stems from the inherent biases of distinct structural patterns across different source domains, which cannot generalize throughout the graph and may obscure truly valuable structural information. For instance, road network data collected from various cities always exhibit structural heterogeneity, where data from some cities may exhibit grid-like structural patterns and data from others have radial configurations \cite{badhrudeen2022geometric}. Such heterogeneity hinders the models from capturing underlying generalizable structural features and therefore limits their performance on real-world applications like traffic flow prediction \cite{zhang2024physics, zhang2025multi} or route planning \cite{zhuang2019toward}. Traditional graph domain adaptation methods rely on training data from different domains and cannot generalize to unseen domains. In contrast, LLMs, with their superior semantic understanding and generalization abilities, can understand different graph structural patterns in a zero-shot manner \cite{wang2024llms}, providing new opportunities for mitigating structural heterogeneity in different graph data from unseen domains.

\textbf{(1) Traditional Methods for Heterogeneous Structure Modeling} While many graph learning methods struggle to generalize across structures from different source domains with different data distributions, recent research investigates how to mitigate this issue by adapting a learned model from a source domain to a target domain. These graph adaptation methods provide solutions for eliminating the biases across different domains and help capture unified and generalizable embeddings for graph structures with excessive heterogeneity. 

Earlier methods for graph domain adaptation, including DANE \cite{zhang2019dane} and ACDNE \cite{shen2020adversarial}, employ shared-weight GNNs to align the embedding spaces of different graphs and utilize a least squares generative adversarial network to regularize the distribution alignment, ensuring that the learned unified embeddings do not include domain-specific information. Different from these adversarial regularization methods, some approaches focus on directly aligning distributions from different domains using different metrics. For example, SR-GNN \cite{zhu2021shift} addresses the domain shifts by regularizing the hidden layer distributions using central moment discrepancy. GDA-SpecReg \cite{you2023graph} combines optimal transport theory and graph filter theory to derive a theoretical bound for graph domain adaptation. Based on this bound, the method utilizes Wasserstein-1 distance to regularize the node representation distributions. StruRW \cite{liu2023structural} proposes an effective approach to reduce conditional structural shifts by re-weighting the edges in the source graph. While most methods align distributions for spatial representations, DASGA \cite{pilanci2020domain} introduces a spectral-based approach by aligning the Fourier bases of the source and target graphs, which ensures that the label functions in both domains have similar coefficients in their respective bases.

While these traditional methods demonstrate superiority in mitigating cross-domain structural heterogeneity, they require separate source and target domain data for training, which limits their application in scenarios without such training data or with a large number of domains. Additionally, traditional GNNs typically rely on fixed k-hop neighborhoods, limiting their ability to generalize across highly diverse connectivity patterns from different source domains with varying hop distances. Furthermore, these methods are inherently dependent on training data, which constrains their ability to handle structures that deviate from the patterns observed during training \cite{zheng2023gnnevaluator}.

\begin{figure}[t]
\centerline{\includegraphics[width=0.9\linewidth]{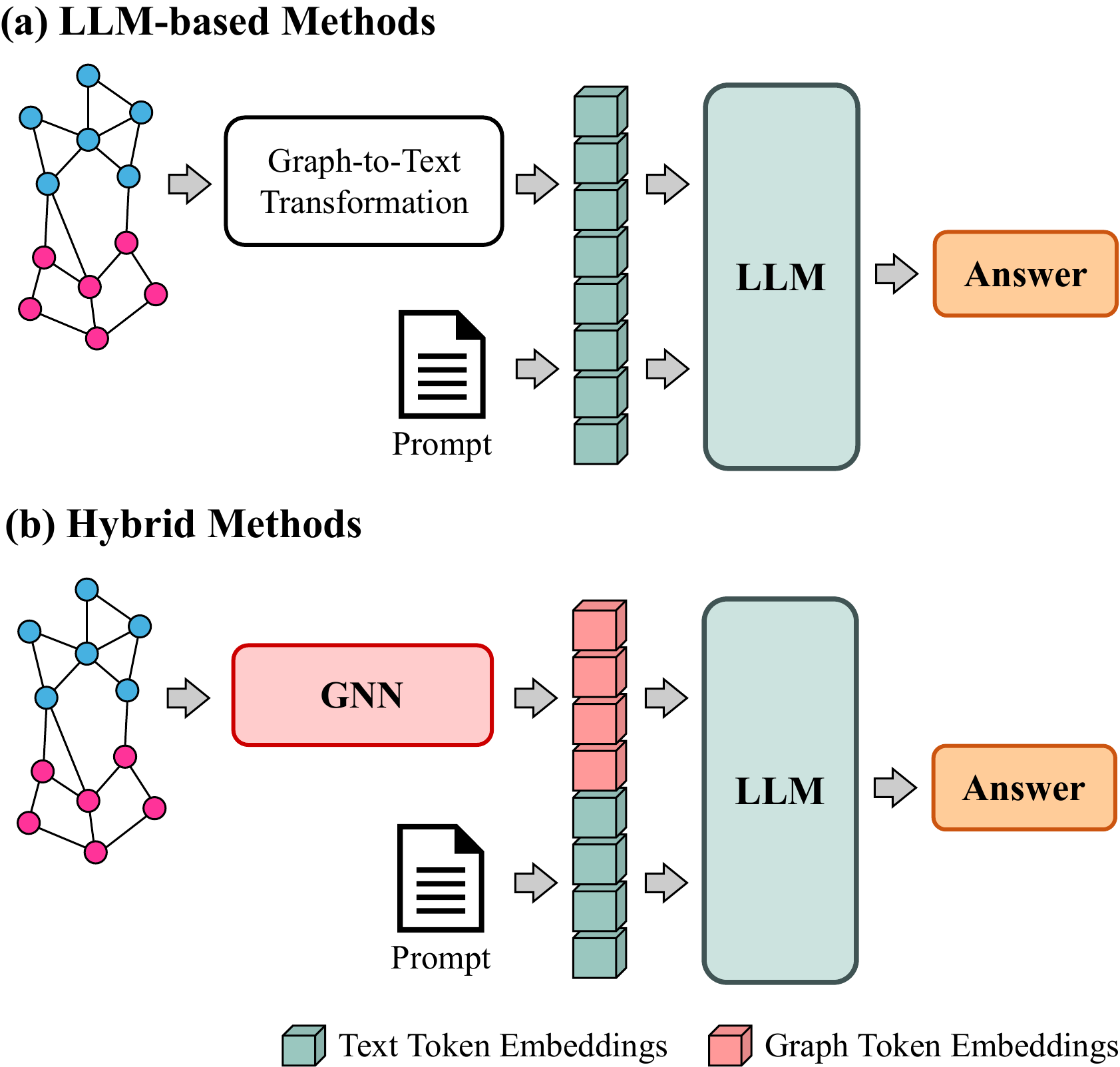}}
\caption{Two different approachs for handling cross-domain structural heterogeneity with LLMs. (a) LLM-based methods convert graph structures into textual descriptions, enabling direct input to LLMs. (b) Hybrid methods align GNN-generated graph token embeddings with the native text token embeddings of LLMs in a unified representation space, thereby enhancing the structural comprehension capabilities of LLMs.}
\label{fig:llm_for_structural_heterogeneity}
\end{figure}

\textbf{(2) LLM-based Methods for Heterogeneous Structure Modeling} 
Different from traditional graph learning methods, LLM-based methods provide a more flexible alternative by representing graph structures through unified text-based descriptions, which are not restricted by a fixed k-hop structural neighborhood assumption. Instead of relying on predefined graph processing mechanisms, LLMs, with superior comprehension and generalization abilities, can understand unseen structural patterns in different domains. Since LLMs are designed to process natural language inputs, a direct way to bridge the gap between graph structures and the input format of LLMs is to transform structural information into textual descriptions, as depicted in Figure~\ref{fig:llm_for_structural_heterogeneity}(a). Here we survey recent advances in structure-to-text transformation methods that enable LLMs to understand heterogeneous structural patterns. Although some methods are not specifically designed for mitigating cross-domain structural heterogeneity, they provide strong potential for addressing the associated challenges.

Recent works have explored how to effectively transform graph structures into text to enable better understanding and reasoning by LLMs. Early methods \cite{liu2023evaluating, wang2023can, fatemi2024talk, guo2023gpt4graph} experimented with existing graph textualization techniques, such as structure description (like node sequences and edge sequences) and formal languages (like graph markup language \cite{brandes2010graph}). While LLMs can comprehend graph structures from these natural language descriptions, their results demonstrate that the choice of different textualization techniques can significantly affect the performance of LLMs on different graph tasks. Different from previous methods that only consider natural language format for representing graph structures, GITA \cite{wei2024gita} transforms graph structures into both text and images to obtain a better understanding of structural information from different modalities, which are then fed into a Vision-Language Model (VLM) for addressing graph tasks. LLM4Hypergraph \cite{feng2025beyond} designs low-order and high-order structure languages to transform hypergraph structures into natural languages. These studies have significantly advanced the exploration of structure-to-text approaches for LLMs to comprehend graph topological structures.

Beyond merely understanding topological structures, textual descriptions can also incorporate the attributes of nodes and edges, providing a richer context that enables LLMs to capture the intricate features of the graph structure. GraphText \cite{zhao2023graphtext} constructs a graph-syntax tree to preserve the hierarchical structure and traverses the tree structure to generate a graph text sequence, where the node attributes are incorporated in the description of leaf nodes. WalkLM \cite{tan2024walklm} generates paths by attributed random walks and textualizes the paths into natural languages for processing in LLMs. To effectively capture cross-group connections while minimizing noisy nodes, Path-LLM \cite{shang2024path} employs shortest paths to generate structural sequences. InstructGLM \cite{ye2024language} describes graph structures using neighbors in different scales and utilizes instruction tuning to finetune an LLM to better perform graph tasks. The textual node attributes are concatenated with the node index to better incorporate semantic information. MuseGraph \cite{tan2024musegraph} incorporates neighbors and paths together to textualize graph structures, which can capture both local connectivity and complex path-based relationships between nodes. Graph-LLM \cite{chen2024exploring} transforms text-attributed graphs into natural languages and demonstrates that LLMs can understand the graph structures from language descriptions and demonstrate remarkable zero-shot performance on graph tasks. 

By converting diverse graph structures into natural language descriptions, these structure-to-text transformation methods enable LLMs to interpret and reason across varying connectivity patterns in a unified manner. These approaches allow LLMs to leverage their pretrained linguistic knowledge to identify valuable relational patterns and facilitate more flexible reasoning across different graph topologies with excessive heterogeneity.

\textbf{(3) Hybrid Methods for Heterogeneous Structure Modeling} While structure-to-text transformation methods offer a straightforward way to bridge the gap between graph structures and the input format of LLMs, they heavily depend on the language comprehension ability of LLMs to correctly interpret the input graph structures. Some recent works explore learning explicit structure representations using traditional graph learning methods like GNNs and aligning these representations with the token space of LLMs, as depicted in Figure~\ref{fig:llm_for_structural_heterogeneity}(b). These hybrid methods leverage the structural modeling capabilities of traditional graph learning methods alongside the comprehension and reasoning strengths of LLMs for more effective graph structure modeling.

Advanced methods bridge the gap between graph structures and LLMs by converting graph data into graph tokens that LLMs can understand. GNP~\cite{tian2024graph} and GraphToken~\cite{perozzi2024let} both use a GNN to encode graph structure and then employ a projector to map these embeddings into the token space of the LLM, allowing them to be processed alongside regular text. LLaGA~\cite{chen2024llaga} takes a different approach by reorganizing graph nodes into structure-aware sequences before projecting them, which helps the model maintain its general-purpose nature. TEA-GLM~\cite{wang2024llms} refines this alignment with feature-wise contrastive learning, using principal components from the token space to precisely map GNN representations. GraphGPT~\cite{tang2024graphgpt} introduces a dual-stage instruction tuning framework that directly aligns graph tokens with LLMs using both self-supervised and task-specific instruction tuning. To generate graph tokens for heterogeneous graphs, HiGPT~\cite{tang2024higpt} introduces an in-context heterogeneous graph tokenizer that encodes diverse node and edge types by using a language-based parameterized heterogeneity projector, dynamically generating graph tokens that represent the heterogeneous semantic relationships.

These hybrid methods incorporate structure representations from traditional graph learning methods to facilitate the structure comprehension ability of LLMs, providing more effective solutions for mitigating the challenges from structural heterogeneity. Additionally, the use of compact graph tokens reduces input size and enables LLMs to process larger structures within their context window, which is crucial for understanding heterogeneous structural patterns across varying scales.

\subsubsection{Evaluation for Handling Cross-Domain Heterogeneity in Graphs}

We summarize the existing evaluation settings for LLM-based methods for handling cross-domain heterogeneity in graphs, covering benchmark datasets, evaluation metrics, and downstream tasks (see Table~\ref{tab:references_heterogeneity_grouped}).

Evaluations typically use well-established traditional graph benchmark datasets. These datasets span a wide range of domains, including citation networks (e.g., Cora~\cite{mccallum2000automating}, Citeseer~\cite{giles1998citeseer}, and PubMed~\cite{sen2008collective}), biomedical graphs (e.g., MIMIC-III~\cite{johnson2016mimic}, MOLHIV~\cite{wu2018moleculenet}, and MOLPCBA~\cite{wu2018moleculenet}), commercial and recommendation networks (e.g., MovieLens~\cite{harper2015movielens}, Netflix~\cite{netflix_prize_data}, and various Amazon product datasets~\cite{zhu2024touchup}), as well as knowledge graphs (e.g., FB15K237~\cite{toutanova2015observed} and WN18RR~\cite{dettmers2018convolutional}). While most of these datasets are standard in graph learning, cross-domain evaluations differ from conventional setups. Instead of training and testing within the same domain, LLM-based methods are often trained on source-domain datasets (or without further training) and tested on unseen target-domain datasets~\cite{zhao2023graphtext, wei2024gita, cheng2024boosting}. In some cases, few-shot settings are also utilized to assess generalization ability when only a limited number of labeled examples in the target domain are available~\cite{he2024unigraph, liu2024one}.

The choice of evaluation metrics typically depends on the task type. Classification tasks adopt Accuracy, Macro-F1, Micro-F1, or AUC, ranking-oriented tasks like recommendation or link prediction use MRR, NDCG, or Hits@N, and generation tasks such as graph-to-text generation rely on BLEU-4, ROUGE-L, CIDEr, METEOR, or CHRF++. Metrics are usually evaluated under zero-shot or few-shot settings to reflect cross-domain generalization. However, traditional metrics may miss cases where outputs are semantically correct but inconsistent in domain-specific format or style. LLMs, in contrast, can capture nuanced correctness across domains based on their strong language comprehension ability, positioning LLM-as-a-Judge~\cite{gu2024survey} a promising direction for future evaluation. The downstream tasks range from node-level and edge-level tasks to graph-level tasks, including node classification, link prediction, and graph classification, as well as more complex settings like recommendation, question answering, section summarization, and graph-to-text generation. This broad coverage sufficiently evaluates the cross-domain generalization ability of LLM-based methods across diverse task types and application scenarios.

\subsubsection{Summary of Heterogeneity} 

LLMs, pretrained on diverse datasets, are inherently capable of integrating information from various sources into a unified semantic space. The reviewed studies show that LLM-integrated models can transfer knowledge across different graph domains without the need for domain-specific retraining. By leveraging the generalizability of LLMs, these models are able to mitigate different forms of cross-domain heterogeneity, including textual attribute heterogeneity, multimodal attribute heterogeneity, and structural heterogeneity. As a result, they demonstrate superior performance in addressing cross-domain heterogeneity, particularly in scenarios where traditional methods are ineffective or inapplicable due to the lack of sufficient training data. Experiments in these papers show improved performance on cross-domain node classification, link prediction, and graph classification tasks when LLMs are used to either encode heterogeneous content or assist in aligning representations.

\subsection{Dynamic Instability in Graphs}

\begin{figure}[th]
\centering
\includegraphics[width=\columnwidth]{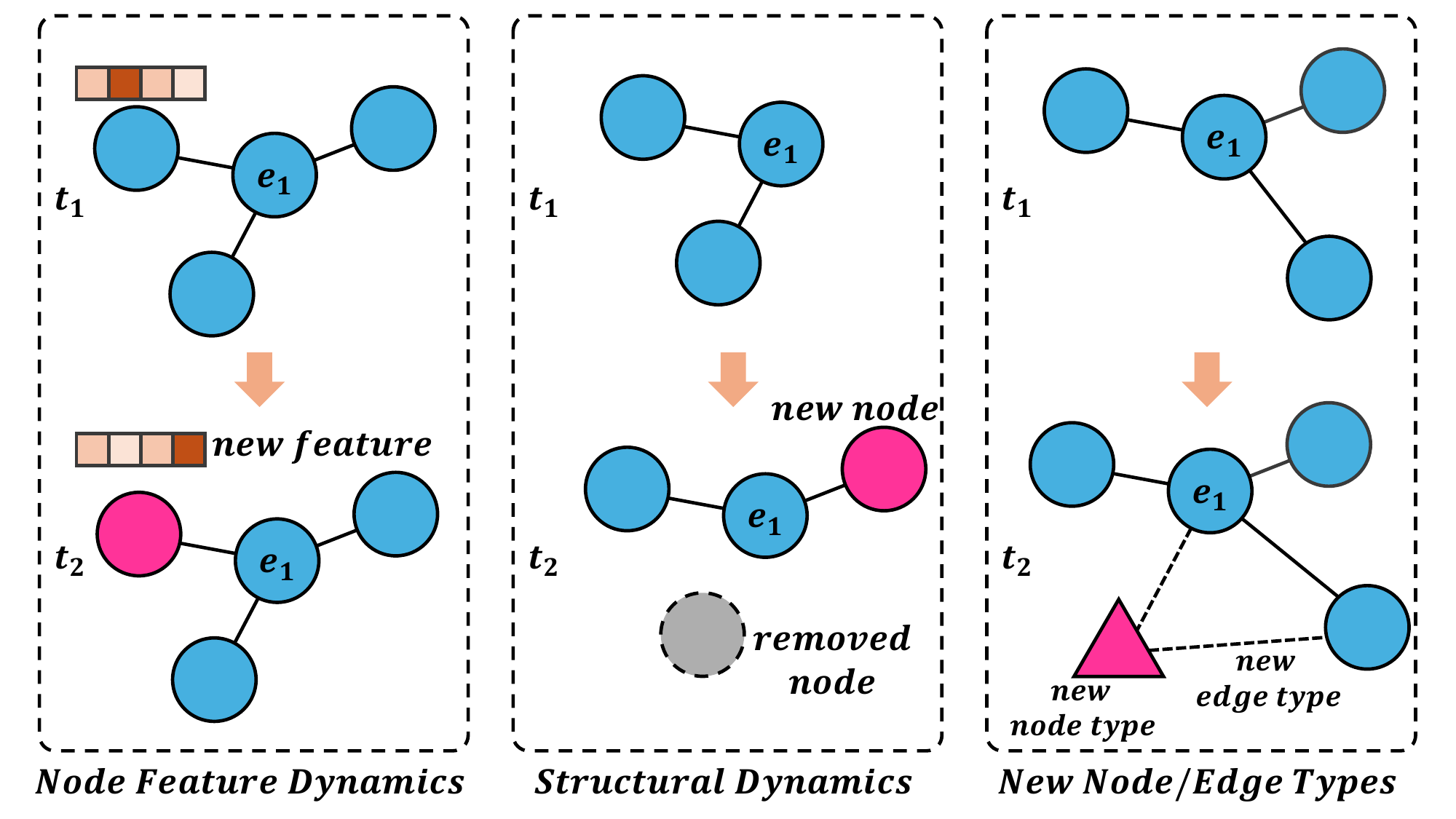}
\caption{Illustration of dynamic instability in graphs. The figure presents three categories of dynamic instability over two timestamps ($t_1$ $\rightarrow$ $t_2$): (1) Node Feature Dynamics, where node attributes change over time; (2) Structural Dynamics, showing newly added node (solid dark purple circle) and newly removed node (gray dashed circle); (3) New Node/Edge Types, depicting the emergence of entirely new node and relationship types, represented by a solid dark purple triangle and dashed edges.}
\label{fig4-1}
\end{figure}

Many real-world systems, from social networks to knowledge bases, are best represented as graphs whose structure and attributes change over time (see Figure ~\ref{fig4-1}). This constant evolution implies a shifting distribution, where patterns observed in the past may not hold true in the future. While gradual evolution poses modeling challenges, the situation becomes significant when the graph undergoes rapid, large-scale, or highly unpredictable changes that alter the graph's characteristics and create significant challenges for model training and inference \cite{9560049}.

Dynamic graphs' instability is challenging to model due to their temporal and structural variability \cite{rossi2020temporalgraphnetworksdeep}. In static settings, models learn stable representations from collected graph data. In contrast, dynamic graphs experience frequent changes in nodes, edges, and attributes. Models struggle to generalize without adaptive mechanisms when faced with structural shifts or abrupt transformations \cite{10.1145/3336191.3371845}. Before the introduction of LLMs, researchers addressed dynamic graphs using snapshot-based and incremental methods. These approaches discretized the graph into time-specific snapshots for independent training \cite{Pareja_Domeniconi_Chen_Ma_Suzumura_Kanezashi_Kaler_Schardl_Leiserson_2020}, or updated models as new nodes and edges appeared \cite{JMLR:v21:19-447}. While effective in some settings, such methods often rely on localized updates and struggle to capture long-range temporal dependencies, particularly in large-scale or rapidly changing graphs.

\begin{figure}[th]
\centering
\includegraphics[width=\columnwidth]{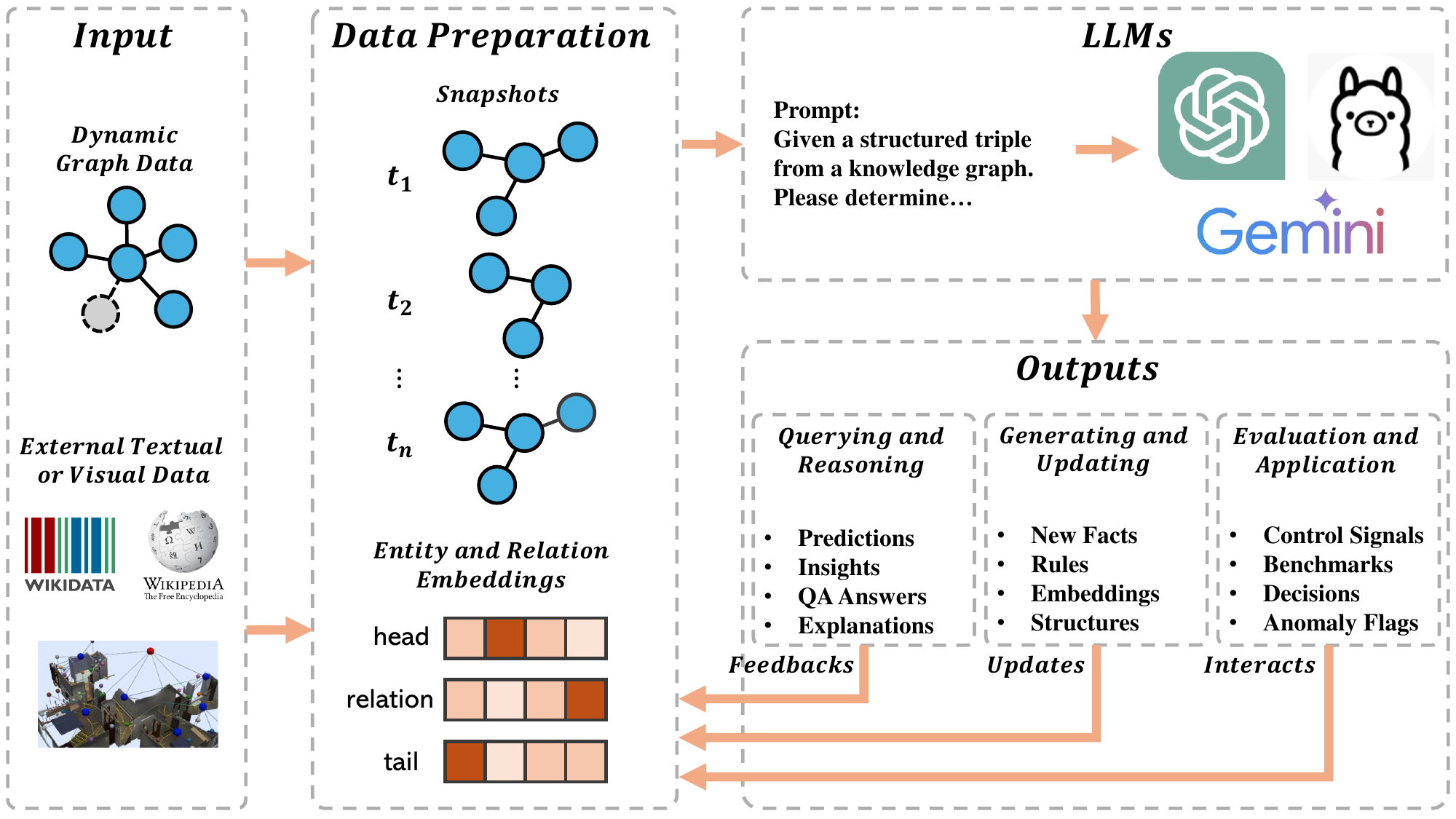}
\caption{A unified framework integrating LLMs for addressing dynamic instability in graphs. Dynamic changes in node features, structure, and new node or edge types are captured by temporal graph embeddings. These dynamic embeddings, combined with context (triples and historical facts), serve as inputs to the LLM. The LLM then performs reasoning tasks and outputs the final prediction or response.}
\label{fig4-2}
\end{figure}

LLMs have recently emerged as a powerful new tool with the potential to address some challenges in dynamic graph learning (Figure~\ref{fig4-2}). By leveraging their strengths in natural language understanding, sequential data processing, few-shot learning, and complex reasoning, LLMs offer novel ways to interpret and model graph dynamics. They can process textual information associated with nodes or edges and understand the semantic context driving structural changes. This infusion of semantic reasoning capabilities opens promising research routes for creating more robust and adaptive models capable of navigating the complexities of dynamic instability. The relevant references and categorization are presented in Table \ref{tab:references_dynamic_instability}.

\begin{table*}[h]\scriptsize
  \centering
  \caption{LLM-based methods for handling \textbf{dynamic instability} in graphs, grouped by tasks, with representative methods, datasets, metrics, and downstream tasks.}
  \renewcommand{\arraystretch}{1.2}
  \setlength{\tabcolsep}{3.5pt}
  \begin{adjustbox}{max width = 1.0\linewidth}
  \begin{tabular}{c P{0.1\linewidth} P{0.1\linewidth} P{0.3\linewidth} P{0.15\linewidth} P{0.15\linewidth}}
    \toprule
    \textbf{Domain} & \textbf{Category} & \textbf{Method} & \textbf{Typical Datasets} & \textbf{Common Metrics} & \textbf{Downstream Tasks} \\
    \midrule

    \multirow{8}{*}{\makecell[c]{Querying \\ and Reasoning}}
    & \multirow{5}{*}{\makecell[c]{Forecasting \\ \& Reasoning}}
      & \textbf{ICL}~\cite{lee-etal-2023-temporal} & WIKI~\cite{10.1145/3184558.3191639}, YAGO~\cite{mahdisoltani:hal-01699874}, ICEWS14~\cite{garcíadurán2018learningsequenceencoderstemporal}, ICEWS18~\cite{jin-etal-2020-recurrent} & MRR, Hits@N & Link Prediction \\
    & & \textbf{zrLLM}~\cite{ding-etal-2024-zrllm} & ICEWS~\cite{DVN/28075_2015}, ACLED~\cite{doi:10.1177/0022343310378914} & MRR, Hits@N & Link Prediction \\
    & & \textbf{CoH}~\cite{xia-etal-2024-chain} & ICEWS14~\cite{garcíadurán2018learningsequenceencoderstemporal}, ICEWS18~\cite{jin-etal-2020-recurrent}, ICEWS05-15~\cite{garcíadurán2018learningsequenceencoderstemporal} & MRR, Hits@N & Link Prediction \\
    & & \textbf{TG-LLM}~\cite{xiong-etal-2024-large} & TGQA~\cite{xiong2024largelanguagemodelslearn}, TimeQA~\cite{chen2021datasetansweringtimesensitivequestions}, TempReason~\cite{tan2023benchmarkingimprovingtemporalreasoning} & F1, Accuracy, Exact Match & Temporal Reasoning \\
    & & \textbf{LLM4DyG}~\cite{10.1145/3637528.3671709} & Enron~\cite{shetty2004enron}, DBLP~\cite{10.1145/1401890.1402008}, Flights~\cite{6846743} & Accuracy, F1, Recall & Spatial-Temporal Reasoning, Graph Reasoning and Querying, Link Prediction \\
    \cmidrule(lr){2-6}
    & \multirow{4}{*}{\makecell[c]{QA \\ \& Interpretability}}
      & \textbf{TimeR4}~\cite{qian-etal-2024-timer4} & MULTITQ~\cite{chen-etal-2023-multi}, TimeQuestions~\cite{10.1145/3459637.3482416} & Hits@N & Temporal Knowledge Graph Question Answering \\
    & & \textbf{GenTKGQA}~\cite{gao-etal-2024-two} & CronQuestion~\cite{saxena2021questionansweringtemporalknowledge}, TimeQuestions~\cite{10.1145/3459637.3482416} & Hits@N & Temporal Knowledge Graph Question Answering \\
    & & \textbf{Unveiling LLMs}~\cite{bronzini2024unveiling} & FEVER~\cite{thorne2018feverlargescaledatasetfact}, CLIMATE-FEVER~\cite{diggelmann2021climatefeverdatasetverificationrealworld} & Precision, Recall, F1, ROC AUC, Accuracy & Claim Verification \\

    \midrule

    \multirow{10}{*}{\makecell[c]{Generating \\ and Updating}}
    & \multirow{6}{*}{\makecell[c]{Generating \\Structures}}
      & \textbf{FinDKG}~\cite{10.1145/3677052.3698603} & WIKI~\cite{10.1145/3184558.3191639}, YAGO~\cite{mahdisoltani:hal-01699874}, ICEWS14~\cite{garcíadurán2018learningsequenceencoderstemporal} & MRR, Hits@N & Link Prediction \\
    & & \textbf{GenTKG}~\cite{liao-etal-2024-gentkg} & ICEWS14~\cite{garcíadurán2018learningsequenceencoderstemporal}, ICEWS18~\cite{jin-etal-2020-recurrent}, GDELT~\cite{leetaru2013gdelt}, YAGO~\cite{mahdisoltani:hal-01699874} & Hits@N & Link Prediction \\
    & & \textbf{Up To Date}~\cite{HATEM2024327} & Wikidata~\cite{vrandevcic2014wikidata} & Accuracy, Response Rate & Fact Validation, Question Answering \\
    & & \textbf{PPT}~\cite{xu-etal-2023-pre} & ICEWS14~\cite{garcíadurán2018learningsequenceencoderstemporal}, ICEWS18~\cite{jin-etal-2020-recurrent}, ICEWS05-15~\cite{garcíadurán2018learningsequenceencoderstemporal} & MRR, Hits@N & Link Prediction \\
    & & \textbf{LLM-DA}~\cite{wang2025large} & ICEWS14~\cite{garcíadurán2018learningsequenceencoderstemporal}, ICEWS05-15~\cite{garcíadurán2018learningsequenceencoderstemporal} & MRR, Hits@N & Link Prediction \\
    \cmidrule(lr){2-6}
    & \multirow{5}{*}{\makecell[c]{Generating Insights \\ \& Representations}}
      & \textbf{TimeLlama}~\cite{10.1145/3589334.3645376} & ICEWS14~\cite{garcíadurán2018learningsequenceencoderstemporal}, ICEWS18~\cite{jin-etal-2020-recurrent}, ICEWS05-15~\cite{garcíadurán2018learningsequenceencoderstemporal} & Precision, Recall, F1, BLEU, ROUGE & Event Forecasting, Explanation Generation \\
    & & \textbf{RealTCD}~\cite{10.1145/3627673.3680042} & Simulation Datasets \cite{10.1145/3627673.3680042} & Structural Hamming Distance, Structural Interventional Distance & Temporal Causal Discovery, Anomaly Detection \\
    & & \textbf{DynLLM}~\cite{zhao2024dynllm} & Tmall~\cite{Tmall}, Alibaba~\cite{ali} & Recall@K, NDCG@K & Dynamic Graph Recommendation, Top-K Recommendation \\

    \midrule

    \multirow{5}{*}{\makecell[c]{Evaluation \\ and Application}}
    & \multirow{3}{*}{\makecell[c]{Model \\Evaluation}}
      & \textbf{Dynamic-TempLAMA}~\cite{margatina-etal-2023-dynamic} & DYNAMICTEMPLAMA~\cite{dhingra2022time} & Accuracy, MRR, ROUGE, F1 & Temporal Robustness Evaluation, Factual Knowledge Probing \\
    & & \textbf{DARG}~\cite{zhang2025darg} & GSM8K~\cite{cobbe2021training}, BBQ~\cite{parrish2021bbq}, BBH Navigate~\cite{suzgun2022challenging}, BBH Dyck Language~\cite{suzgun2022challenging} & Accuracy, Complexity-Induced Accuracy Retention Rate, Exact Match, Accuracy & Mathematical Reasoning, Social Reasoning, Spatial Reasoning, Symbolic Reasoning \\
    \cmidrule(lr){2-6}
    & \multirow{3}{*}{\makecell[c]{Downstream \\Applications}}
      & \textbf{AnomalyLLM}~\cite{liu2024anomalyllm} & UCI Messages~\cite{opsahl2009clustering}, Blogcatalog~\cite{tang2009relational} & AUC & Anomaly Detection \\
    & & \textbf{MoMa-LLM}~\cite{honerkamp2024language} & iGibson scenes~\cite{li2021igibson} & AUC, Recall & Semantic Interactive Object Search \\
    & & \textbf{TRR}~\cite{koa2024temporal} & Reuters Financial News~\cite{ding2014using} & AUROC & Event Detection \\
    
    \bottomrule
  \end{tabular}
  \end{adjustbox}
  \label{tab:references_dynamic_instability}
\end{table*}

\subsubsection{Querying and Reasoning in Dynamic Graphs}

This category includes work that uses LLMs to query, reason about, or retrieve knowledge from dynamic graphs. These approaches analyze or validate existing graph structures rather than actively generating or modifying node and edge features. LLMs act as information retrieval or reasoning tools, querying, analyzing, or inferring over existing graph structures and knowledge to discover new relationships or validate known ones. The key aspect is that LLMs assist in analyzing and understanding the graph's dynamic evolution.

\textbf{(1) Traditional Methods}
Before the emergence of LLMs, querying and reasoning on dynamic graphs relied on specialized graph representation learning and temporal modeling methods. For example, Temporal Graph Networks (TGNs) \cite{rossi2020temporalgraphnetworksdeep}, and related methods use memory modules combined with graph neural networks to process continuous-time dynamic graphs and capture node evolution. Methods like DyRep \cite{osti_10190736} and TGAT \cite{tgat_iclr20} focus on learning node representations that reflect temporal changes in neighborhood structures and interaction patterns. For reasoning on Temporal Knowledge Graphs (TKGs), such as link prediction, traditional approaches employed embedding techniques or rule-based systems to capture changes in entities and relations over time \cite{su2024temporalknowledgegraphquestion}. While effective for specific tasks, these methods struggle to fully use rich semantic information and face challenges in handling zero-shot relations or performing complex multi-step reasoning \cite{10.24963/ijcai.2023/734}.

\textbf{(2) Forecasting and Reasoning}
Current research now explores the semantic understanding and reasoning capabilities of LLMs to address querying and reasoning challenges in dynamic graphs, particularly for TKG forecasting and reasoning. One surprising finding is that LLMs, even without fine-tuning, can achieve performance comparable to specialized TKG models on forecasting tasks simply by using In-Context Learning (ICL) with historical facts converted to text. Their performance holds even when entity names are replaced with numerical IDs, suggesting LLMs can leverage structural and temporal patterns in the context \cite{lee-etal-2023-temporal}. To deepen LLM reasoning, CoH (Chain-of-History) \cite{xia-etal-2024-chain} proposes a method to explore high-order historical information step-by-step, overcoming the limitation of relying only on first-order history and improving temporal reasoning, especially as a plug-in module for graph-based models. TG-LLM \cite{xiong-etal-2024-large} trains an LLM to translate text context into a latent temporal graph and then uses Chain-of-Thought (CoT) reasoning over this graph, enhancing generalizable temporal reasoning. To address the challenge of unseen relations in TKGs, zrLLM \cite{ding-etal-2024-zrllm} uses LLMs to generate semantic representations from relation descriptions, enabling embedding models to recognize zero-shot relations via semantic similarity. Furthermore, LLM4DyG \cite{10.1145/3637528.3671709} introduces a systematic evaluation of LLMs' spatial-temporal understanding on general dynamic graphs. It introduces a benchmark and proposes the ``Disentangled Spatial Temporal Thoughts'' prompting method to improve performance on tasks like link prediction and node classification, although challenges remain with large or dense dynamic graphs.

\textbf{(3) QA and Interpretability}
Another line of research focuses on using LLMs for more complex information retrieval from dynamic graphs, such as Temporal Knowledge Graph Question Answering (TKGQA), and understanding the LLMs' own reasoning processes. TKGQA requires understanding complex temporal constraints in questions and retrieving answers from dynamic knowledge, a task where traditional methods struggle with semantics. TimeR4 \cite{qian-etal-2024-timer4} introduces a time-aware Retrieve-Rewrite-Retrieve-Rerank framework. It uses LLMs and retrieved TKG facts to handle time constraints and reduce temporal hallucination by rewriting questions and reranking retrieved facts. Similarly, GenTKGQA \cite{gao-etal-2024-two} employs a two-stage approach: the LLM first mines constraints to guide subgraph retrieval, then generates answers by fusing GNN signals with text representations via instruction tuning. These methods demonstrate the potential of retrieval-augmented LLMs for complex dynamic knowledge QA. Additionally, to understand how LLMs process factual knowledge, Unveiling LLMs \cite{bronzini2024unveiling} uses dynamic knowledge graphs as a tool to interpret LLM reasoning. It decodes internal token representations layer-wise during fact verification, revealing how factual representations evolve within the LLM.

\subsubsection{Generating and Updating in Dynamic Graphs}

This category includes papers that use LLMs to actively generate new nodes, edges, or their features or to update dynamic graph structures. These approaches typically serve to complete, reconstruct, or adapt the graph structure to dynamic changes. LLMs function as generators or updaters, creating new node/edge attributes or descriptions to fill in missing information or updating the graph structure in real time based on changes. The focus is on utilizing LLMs' generative capabilities to actively create or modify the graph structure to cope with dynamics.

\textbf{(1) Traditional Methods}
Before LLMs became widespread, addressing dynamic graph generation and updates relied mainly on graph generative models and dynamic embedding techniques. Models like GraphRNN \cite{you2018graphrnn} and GRAN \cite{10.5555/3454287.3454670} focused on generating static graph structures; extending them to dynamic settings proved challenging. For updating representations in dynamic graphs, methods like DyGEM \cite{goyal2018dyngem} adapted node embeddings at each time step to reflect graph changes. Knowledge graph updates often depended on manual editing, rule-based systems, or specific database maintenance procedures, which were difficult to automate and could not respond rapidly to real-world changes. These traditional methods generally lacked the ability to use external unstructured information (like text) to guide updates and had limitations in generating semantically consistent and structurally complex graph evolution patterns.

\textbf{(2) Generating Structures}
Researchers now use LLMs' generative power to create or maintain evolving knowledge structures. The FinDKG \cite{10.1145/3677052.3698603} demonstrates the potential of LLMs as dynamic knowledge graph (DKG) generators; their ICKG model builds a DKG directly from financial news text to capture market trends. To address the issue of outdated information in knowledge graphs, Up To Date \cite{HATEM2024327} proposes a method combining LLM reasoning and RAG. It automatically identifies potentially outdated facts in a KG and retrieves information from trusted sources to generate accurate corrections, enabling automated KG maintenance. For TKG completion and forecasting, PPT \cite{xu-etal-2023-pre} converts TKG facts and time intervals into prompted natural language sequences, using a masked language modeling task for completion. GenTKG \cite{liao-etal-2024-gentkg} also adopts a generative forecasting approach for TKGs. It uses a retrieval-augmented framework with parameter-efficient instruction tuning to generate future facts, achieving strong performance and generalization even with minimal training data. Furthermore, LLM-DA \cite{wang2025large} innovatively uses an LLM to generate temporal logical rules from historical TKG data to guide reasoning. A dynamic adaptation strategy updates these rules based on new events, allowing the model to adapt to knowledge evolution without retraining the LLM.

\textbf{(3) Generating Insights and Representations}
Beyond directly generating graph structures or facts, LLMs generate higher-level information like explanations, causal hypotheses, or enhanced node representations to aid dynamic graph analysis and applications. TimeLlama \cite{10.1145/3589334.3645376} focuses on explainable temporal reasoning. It not only predicts future events but also uses an LLM to generate natural language explanations based on historical TKG paths, increasing model trustworthiness. To explore the underlying mechanisms of dynamic systems, RealTCD \cite{10.1145/3627673.3680042} employs LLMs to process textual information (\emph{e.g.}, system logs) and integrate domain knowledge. Through LLM-guided meta-initialization, it improves the quality of temporal causal discovery, especially in industrial scenarios lacking intervention targets. For dynamic recommendation systems, DynLLM \cite{zhao2024dynllm} uses LLMs to generate multi-dimensional user profiles (\emph{e.g.}, interests, preferred brands) and their embeddings from the textual features of purchase histories. These LLM-generated, dynamically updated user profiles enrich user node information, improving the recommendation system's ability to adapt to changing user preferences. These studies show that LLMs can produce basic graph elements and high-level semantic information, offering new ways to understand and model the complex evolution of dynamic graphs.

\subsubsection{Evaluation and Application in Dynamic Graphs}

This category includes work that uses LLMs to evaluate the effectiveness of dynamic graph learning models or apply LLMs in downstream tasks such as link prediction, node classification, or recommendation systems. LLMs serve as evaluation tools or modules within downstream applications, assessing the performance of dynamic graph models or providing additional guidance during training or inference. The emphasis is on using LLMs' comparative, assessment, or decision-support capabilities to handle tasks in dynamic environments.

\textbf{(1) Traditional Methods}
Before using LLMs in these tasks, researchers relied on models that predicted future links or node states based on historical graph data \cite{jin-etal-2020-recurrent}. However, standard evaluation metrics often fail to capture the predictions' semantic quality or real-world plausibility \cite{xiong2025surveylinkpredictiontemporal}. These metrics also struggled to assess how models handled unexpected structural changes or evolving concepts. In downstream applications, tasks such as dynamic anomaly detection typically use statistical methods to identify shifts in graph structure or connectivity patterns \cite{10.1145/3018661.3018731}. Dynamic recommendation systems often relied on sequential models that analyzed user interaction histories \cite{DBLP:journals/corr/HidasiKBT15} or time-aware collaborative filtering. Although these approaches were effective for specific problems, they lacked the ability to perform deeper semantic reasoning.

\textbf{(2) Model Evaluation}
LLMs provide new perspectives for evaluating model performance and robustness in dynamic settings. Dynamic-TempLAMA \cite{margatina-etal-2023-dynamic} presents a dynamic benchmarking framework specifically designed to assess how well pretrained language models (MLMs) handle temporal concept drift—the evolution of factual knowledge over time. The framework dynamically creates time-sensitive test sets from Wikidata and evaluates MLMs through multiple views (probing, generation, scoring) to determine if their internal knowledge is outdated. DARG \cite{zhang2025darg} addresses the limitations of static benchmarks by proposing a method to dynamically generate new evaluation data. It introduces changes to the reasoning graphs of existing benchmark samples to create novel test data with controlled complexity and diversity, using a code-augmented LLM to ensure label correctness. This enables adaptive evaluation of LLMs' reasoning capabilities as they evolve. Both works highlight the importance of moving beyond static, snapshot-based evaluations toward more dynamic, adaptive approaches, especially as both models and world knowledge constantly change.

\textbf{(3) Downstream Applications}
LLMs are also embedded directly into downstream applications that process dynamic graph data, serving as core reasoning or decision-making components. AnomalyLLM \cite{liu2024anomalyllm} uses LLM knowledge for few-shot anomaly edge detection in dynamic graphs. By aligning edge representations with word embedding prototypes and using in-context learning, the method effectively identifies emerging anomaly types with few labeled examples, demonstrating LLMs' potential for adapting to changing environments like cybersecurity. In robotics, MoMa-LLM \cite{honerkamp2024language} integrates an LLM with dynamically updated, open-vocabulary scene graphs representing an explored environment. The LLM reasons over this evolving graph to guide a mobile robot in long-horizon, interactive object search tasks, showcasing how LLMs can integrate dynamic spatial-semantic information for high-level planning. The financial sector also leverages LLMs; TRR (Temporal Relational Reasoning) \cite{koa2024temporal} uses an LLM-based framework mimicking human cognition (memory, attention) to detect potential stock portfolio crashes. It reasons over dynamically generated temporal relational information extracted from news to assess the aggregated impact of evolving events, which is useful for rare events lacking historical data. These applications demonstrate LLMs acting as powerful reasoning engines in complex real-world tasks that require understanding and responding to dynamically changing structured information.

\subsubsection{Evaluation for Handling Dynamic Instability in Graphs}

We review the evaluation pipeline of LLM-based methods for addressing \emph{dynamic instability} in graphs, covering benchmark datasets, evaluation metrics, and downstream tasks (see Table~\ref{tab:references_dynamic_instability}). Commonly used datasets include temporal knowledge graph benchmarks such as ICEWS14~\cite{garcíadurán2018learningsequenceencoderstemporal}, ICEWS18~\cite{jin-etal-2020-recurrent}, ICEWS05-15~\cite{garcíadurán2018learningsequenceencoderstemporal}, YAGO~\cite{mahdisoltani:hal-01699874}, and WIKI~\cite{10.1145/3184558.3191639} for link prediction and temporal reasoning. Event-based corpora such as ACLED~\cite{doi:10.1177/0022343310378914}, GDELT~\cite{leetaru2013gdelt}, and TempReason~\cite{tan2023benchmarkingimprovingtemporalreasoning} are used for temporal forecasting, while domain-specific datasets like Enron~\cite{shetty2004enron}, DBLP~\cite{10.1145/1401890.1402008}, Tmall~\cite{Tmall}, and Alibaba~\cite{ali} support communication analysis and recommendation. Temporal KGQA relies on MULTITQ~\cite{chen-etal-2023-multi}, TimeQuestions~\cite{10.1145/3459637.3482416}, and CronQuestion~\cite{saxena2021questionansweringtemporalknowledge}, and claim verification uses FEVER~\cite{thorne2018feverlargescaledatasetfact} and CLIMATE-FEVER~\cite{diggelmann2021climatefeverdatasetverificationrealworld}. Anomaly and event detection adopt UCI Messages, Blogcatalog, iGibson scenes, and Reuters Financial News.

Evaluation metrics vary by task: MRR and Hits@N dominate ranking tasks; Accuracy, F1, Precision, Recall, and ROC-AUC are applied in classification; BLEU, ROUGE, and Exact Match assess generative tasks; Recall@K and NDCG@K measure recommendation; and SHD or SID evaluate temporal causal discovery. AUC or AUROC are common for anomaly detection.

The surveyed methods span diverse downstream tasks, including link prediction, temporal reasoning, temporal KGQA, claim verification, fact validation, dynamic graph recommendation, temporal causal discovery, event forecasting, and anomaly or event detection. This diversity highlights the capability of LLMs to integrate temporal reasoning and external knowledge for robust performance in evolving graph environments.

\subsubsection{Summary of Dynamic Instability}

The integration of LLMs with dynamic graph learning presents a promising direction for addressing the inherent challenges posed by evolving graph structures. Leveraging their powerful sequence modeling capabilities and the flexibility of natural language encoding, LLMs can effectively adapt to distribution shifts and temporal variations that traditional static graph models struggle to capture. Moreover, LLMs can incorporate external temporal knowledge into the graph reasoning process, enhancing predictive power and enabling anticipation of future changes rather than merely reacting to them. Early empirical evidence suggests that LLM-augmented dynamic graph models exhibit greater robustness and sustained performance over time. Nevertheless, these benefits come with increased computational costs and complexity, requiring careful architectural design and error mitigation strategies to prevent error propagation and maintain reliability in long-term deployment.

\section{Limitations, Challenges, and Future Directions}\label{s4}

This section builds upon the preceding analysis to consolidate key challenges and limitations in graph-based learning with LLMs, and to outline potential directions for future research. The discussion is organized around several considerations, including \textbf{efficiency and scalability}, \textbf{explainability, fairness, and faithfulness}, and \textbf{security, robustness, and governance}, which together reflect both the current barriers and the broader goals for advancing LLM–graph integration.

\subsection{Efficiency and Scalability}
Scalability is a key challenge in integrating LLMs with graph data, given the high memory and computation costs of large, heterogeneous, and dynamic graphs. We summarize efficiency challenges across scenarios and explore ways to improve tractability, compactness, and inference speed.

\subsubsection{Incompleteness}
LLM-based methods for graph completion show promise in generating missing nodes, edges, or attributes, but they also introduce risks of hallucination and structural inconsistency. To improve the fidelity of generated content, future research should explore structure-constrained decoding, confidence-aware integration, and post-hoc validation using GNNs or external knowledge bases \cite{sehwag2024context}. Incorporating structural priors from graphs directly into training objectives may further enhance the alignment between textual reasoning and graph topology \cite{zhang2024making}. At the same time, reducing representational redundancy is critical for scalability, motivating the development of compact graph–text co-representations. For example, injecting structure tokens or refactoring sequence layouts can reduce the need for full-graph serialization during inference \cite{xue2024leading,coppolillo2025injectkg}. Besides, efficiency-oriented designs should also consider concrete resource metrics such as sequence length, memory usage, and per-sample or per-token computation cost \cite{fan2024graphllm_survey}.

\subsubsection{Imbalance}
Imbalanced distributions are common in real-world graphs, where a few dominant classes overshadow long-tail entities that often lack sufficient structural and semantic support \cite{ko2024subgraph,guo2024graphedit}. Uniformly applying LLMs across all samples is inefficient and may overfit majority patterns while underperforming on minority nodes. Moreover, hallucinations in underrepresented regions pose reliability risks, especially when topological signals are weak \cite{zhang2024can}. To mitigate these issues, future research should consider selective LLM invocation for hard or minority-class cases, while using GNNs for routine instances and transferring LLM knowledge into lightweight models via parameter-efficient tuning \cite{hu2021lora,dettmers2023qlora}. Structure-aware validation modules that cross-check LLM outputs against graph-derived patterns can improve robustness \cite{wei2023kicgpt}, and a promising direction is to construct joint text–graph causal representations to help models infer the semantic origins of structural sparsity and synthesize logically grounded virtual connections \cite{xu2024multi}.

\subsubsection{Cross-Domain Heterogeneity}
Graphs that span multiple domains or modalities often exhibit heterogeneity in node types, attribute formats, and structural patterns, which increases prompt length and introduces misalignment between textual and topological inputs \cite{he2024harnessing,cui2024multimodal}. Existing methods struggle to handle these inconsistencies, as they typically rely on shallow normalization or isolated attribute encoding, and LLMs themselves lack permutation invariance when processing graph structures \cite{zhao2023graphtext}. To address these limitations, future efforts should explore context-aware modeling that jointly encodes multimodal attributes, local connectivity, and task-specific signals. Symmetry-preserving techniques such as contrastive learning or explicit prompt designs can help models distinguish meaningful structural variations from order artifacts \cite{ye2024language}. Additionally, tighter LLM–GNN integration—via dual encoders or knowledge distillation from graph-level GNNs—may provide a more balanced trade-off between semantic richness and structural fidelity \cite{tian2024graph,perozzi2024let}.

\subsubsection{Dynamic Instability}
In dynamic graphs, frequent structural or attribute updates introduce substantial overhead for model recomputation and pose challenges for maintaining temporal consistency \cite{lee-etal-2023-temporal}. Current LLM-based approaches often rely on full re-encoding or static snapshots, which fail to reflect evolving semantics and result in stale or inaccurate reasoning \cite{xiong-etal-2024-large}. To overcome these limitations, future research should incorporate time-aware prompting, localized subgraph updates, and selective verification to reduce redundant computation \cite{fan2024graphllm_survey}. Caching frequently accessed substructures or intermediate representations can further improve efficiency. More importantly, integrating GNNs with LLMs for temporal causal reasoning, along with pretraining strategies that embed temporal priors, may help models adapt to semantic drift and support robust inference over evolving graph states \cite{xia-etal-2024-chain,ding-etal-2024-zrllm}.

\subsection{Explainability, Fairness, and Faithfulness}

LLM-based graph systems pose serious challenges in explainability, fairness, and faithful reasoning—especially in scenarios involving incomplete data, long-tail distributions, domain shifts, and temporal updates. While current approaches offer partial solutions, their limitations in interpretability, causal alignment, and robustness across scenarios call for more principled advancements.

\subsubsection{Incompleteness}
Incomplete graphs often result in LLM outputs that lack faithfulness, as the underlying reasoning is not grounded in observable evidence \cite{chen2023knowledge,xu2024generate,sehwag2024context}. This issue is exacerbated by decoding processes that fail to enforce topological constraints and by the absence of systematic post-generation verification. Future work should incorporate structure-constrained decoding objectives and apply graph consistency losses during training \cite{zhang2024making}, while complementing generation with post-hoc validation using GNN-based scorers or confidence filters \cite{chai2023graphllm,wang2024llm}. Furthermore, reasoning fidelity deteriorates significantly when only partial graph observations are available \cite{xu2024generate}. To address this, prompting strategies should explicitly encode subgraph structures to help LLMs reason over local topological contexts. In addition, iterative refinement mechanisms that alternate between retrieval, reasoning, and validation can simulate GNN-style aggregation while supporting auditable evidence chains.

\subsubsection{Imbalance}  
Graph imbalance not only degrades predictive performance but also undermines explanation stability and reliability. Due to sparse semantics and weak connectivity, LLMs may struggle to generate faithful rationales for long-tail nodes. Moreover, existing interpretability tools rarely account for such imbalance, leading to inconsistent explanations under input perturbations. Future work should explore causal mechanisms underlying topological sparsity and use LLMs to generate logically consistent virtual subgraphs to restore class-level balance \cite{guo2024graphedit}. At the same time, interpretability evaluation should emphasize stability: explanations for minority-class nodes should remain consistent across counterfactual edits or input augmentations, and validation modules should reject rationales that do not correspond to truly discriminative substructures.

\subsubsection{Cross-Domain Heterogeneity}
Inconsistencies in label semantics, graph patterns, and textual attributes across domains make it difficult for LLMs to produce generalizable explanations. These mismatches often introduce domain-specific biases that harm fairness and reasoning quality \cite{dong2023fair_graph_mining,gallegos2023bias_llm}. Current models lack the ability to maintain explanation coherence when transferred across domains \cite{ying2019gnnexplainer,lucic2022cfg}. Future systems should develop domain-invariant rationale templates and establish multi-level auditing frameworks covering embedding spaces, output distributions, and textual justifications to detect and mitigate explanation drift. Additionally, aligning the structural reasoning paths produced by GNNs with the textual rationales generated by LLMs can bridge the gap between semantics and topology and enhance explanation faithfulness.

\subsubsection{Dynamic Instability}
The evolving nature of real-world graphs introduces temporal variations that challenge the stability and trustworthiness of model explanations. As nodes and edges change over time, previously valid rationales may become outdated or misleading, yet current systems lack temporal auditing mechanisms. Future work should introduce rolling, time-aware auditing protocols that explicitly link model outputs with graph update events \cite{dong2023fair_graph_mining}. Moreover, enabling executable explanations—where LLM-generated rationales are translated into symbolic constraints or verifiable subgraph traces—can improve the reliability and traceability of reasoning under temporal drift \cite{lyu2023faithfulcot}.

\subsection{Security, Robustness, and Governance}

As graph–LLM systems become increasingly applied in high-stakes domains such as finance, healthcare, and recommendation, they face growing vulnerabilities arising from incomplete information, class imbalance, cross-domain inconsistency, and evolving graph structures. These challenges expose the system to risks including data poisoning, prompt injection, structure manipulation, and adversarial generalization, many of which are amplified by the flexible yet opaque behavior of LLMs.

\subsubsection{Incompleteness}
In settings where graph data is sparse or noisy, even minor perturbations to node or edge attributes can lead to significant shifts in completion outcomes, compromising the integrity of the generated graph \cite{zugner2018nettack,zhang2019poisonkg}. Similarly, LLM-augmented retrieval systems are vulnerable to prompt-level attacks, where indirect injection or document poisoning misleads the model's response \cite{stefano2024rag_indirect}. To improve robustness, future work should investigate multi-layer defenses, including data-level sanitization and edge pruning, model-level protection through graph denoising techniques such as GNNGuard \cite{zhang2020gnnguard}, and certification-based methods like randomized smoothing to quantify the system’s tolerance under adversarial conditions \cite{bojchevski2019cert,wang2021certified}.

\subsubsection{Imbalance}  
Nodes associated with rare categories or low connectivity often lack sufficient structural context, making them particularly susceptible to adversarial influence; subtle perturbations can alter neighborhood aggregation or trigger class boundary shifts \cite{jin2020review}. Moreover, few-shot trigger constructions may induce undesirable distributional shifts in LLM outputs. To address these issues, future systems should adopt targeted defenses for long-tail nodes, including structural consistency checks and trigger detection via adversarial training regimes that expose the model to rare-case manipulations.

\subsubsection{Cross-Domain Heterogeneity}
When LLM-based models are deployed across domains with differing label semantics, structure formats, or template designs, inconsistencies can be exploited to induce behavior shifts or prompt corruption. Existing systems often lack robust controls over prompt templates or graph encodings, making them vulnerable to injection and escalation attacks \cite{stefano2024rag_indirect}. Looking forward, stronger domain-level governance is needed, including adversarial evaluation prior to deployment, enforcement of signed prompt templates, and whitelist-based control over structural tokens to prevent unauthorized modifications.

\subsubsection{Dynamic Instability}
Dynamic graphs introduce further complexity, as attackers may inject or remove connections during critical time windows or log adversarial traces to shape long-term model behavior \cite{jin2020review}. These time-sensitive attacks are often stealthy and hard to reverse, leading to compounding damage \cite{wang2021certified}. Future directions should prioritize rolling-window detection and temporal rollback mechanisms, coupled with causal tracing of performance degradation to structural changes. Additional safeguards such as time-aware scoring and path-verifiable explanation generation can help separate legitimate evolution from malicious interference.

\section{Conclusion} \label{s5}

This survey has presented a comprehensive and systematic review of how recent advances in LLMs can be leveraged to address four fundamental, data-centric challenges in graph learning: incompleteness of structures or attributes, severe imbalance in node and edge distributions, cross-domain heterogeneity in semantics and structure, and dynamic instability arising from evolving topologies and interactions. To achieve this, we conducted an extensive literature collection and categorization, and organized representative methods. We further summarized benchmark datasets and evaluation metrics used in existing studies, assessed empirical trends. In addition, we identified open technical challenges and outlined promising future research directions.

We have synthesized a broad spectrum of LLM–graph integration strategies under this framework, demonstrating how LLMs bring distinctive capabilities that complement purely graph-based approaches. In the context of incompleteness, LLMs apply semantic reasoning and draw upon external knowledge to infer missing attributes and relationships, serving as intelligent imputers. For imbalanced graphs, they can generate synthetic samples and enrich feature spaces, thereby enhancing minority-class representations and mitigating bias. In heterogeneous graph scenarios, LLMs facilitate the unification of disparate modalities and domain-specific schemas into coherent embeddings, enabling effective cross-domain alignment. For dynamic graphs, their contextual and temporal reasoning allows for anticipating structural changes, explaining evolution, and supporting continuous adaptation.

Looking ahead, the integration of LLMs with graph learning still faces important open questions. Improving efficiency and scalability is essential for large-scale or real-time applications, while enhancing interpretability and trustworthiness will be critical for deployment in high-stakes domains such as healthcare and finance. Developing mechanisms for continual adaptation without catastrophic forgetting remains a significant challenge, as does bridging the gap between textual knowledge encoded in LLMs and structural graph signals. In addition, controlling hallucination in LLM-generated graph content is vital to ensure both semantic validity and structural consistency.

In conclusion, the convergence of LLMs and graph learning marks a promising new direction, combining the deep semantic understanding of natural language processing with the structured relational modeling of graph machine learning. This synergy has already yielded models that are more robust, knowledgeable, and adaptable than those based on either technology alone. As research advances, we anticipate rapid progress toward graph learning systems with greater generality and intelligence, capable of reasoning effectively over the rich, dynamic networks that underpin real-world data.

\section*{Author Contributions}
\textbf{Mengran Li:} Conceptualization, Methodology, Formal analysis, Writing – original draft, Visualization.
\textbf{Pengyu Zhang:} Conceptualization, Methodology, Formal analysis, Writing – original draft, Visualization.
\textbf{Wenbin Xing:} Conceptualization, Data curation, Investigation, Writing – original draft, Visualization.
\textbf{Yijia Zheng:} Conceptualization, Data curation, Investigation, Writing – original draft, Visualization.
\textbf{Klim Zaporojets:} Software, Validation, Writing – review \& editing.
\textbf{Junzhou Chen:} Supervision, Writing – review \& editing.
\textbf{Ronghui Zhang:} Supervision, Project administration, Funding acquisition, Writing – review \& editing.
\textbf{Yong Zhang:} Supervision, Resources.
\textbf{Siyuan Gong:} Supervision.
\textbf{Jia Hu:} Supervision, Resources.
\textbf{Xiaolei Ma:} Supervision, Resources.
\textbf{Zhiyuan Liu:} Supervision, Writing – review \& editing.
\textbf{Paul Groth:} Supervision, Writing – review \& editing.
\textbf{Marcel Worring:} Supervision, Writing – review \& editing.

\section*{Declaration of competing interest}
The authors declare that they have no known competing financial interests or personal relationships that could have appeared to influence the work reported in this paper.

\section*{Acknowledgements}
This work was supported in part by the Shenzhen Science and Technology Program under Grant JCYJ20240813151129038; in part by the National Natural Science Foundation of China under Grant 52172350, Grant 51775565, and Grant W2421069; in part by the Guangdong Basic and Applied Research Foundation under Grant 2022B1515120072; and in part by the Science and Technology Planning Project of Guangdong Province under Grant 2023B1212060029.

\bibliographystyle{elsarticle-num} 
\bibliography{bib}

\end{document}